\documentclass{article}

\usepackage{arxiv}

\usepackage[utf8]{inputenc} 
\usepackage[T1]{fontenc}    
\usepackage{amsmath}
\usepackage{hyperref}       
\usepackage{url}            
\usepackage{booktabs}       
\usepackage{amsfonts}       
\usepackage{nicefrac}       
\usepackage{microtype}      
\usepackage{lipsum}		
\usepackage{graphicx}
\usepackage{epstopdf}
\usepackage{natbib}
\usepackage{doi}
\usepackage{amssymb}
\usepackage{amsmath}
\usepackage{algorithm}
\usepackage{algorithmic}
\usepackage{mathtools}
\usepackage{enumerate}
\usepackage{subfig}
\usepackage{caption}
\usepackage{multirow}
\usepackage{multicol}
\usepackage{stackengine}
\usepackage{nicefrac}       
\usepackage{Definitions}

\DeclareSymbolFont{bbold}{U}{bbold}{m}{n}
\DeclareSymbolFontAlphabet{\mathbbold}{bbold}

\DeclareGraphicsExtensions{.pdf,.jpeg,.png,.eps}
\renewcommand{\shorttitle}{Explaining Decisions of Neural Classifiers Based on Set Membership Estimation}

\title{A Set Membership Approach to Discovering Feature Relevance and Explaining Neural Classifier Decisions}

\author{Stavros P.~Adam$^{1}$, ~Aristidis C.~Likas$^{2}$, \\
  $^{1}$ {Department of Informatics and Telecommunications, University of Ioannina, Arta, Greece} \\ $^{2}$ {Department of Computer Science and Engineering, University of Ioannina, Ioannina, Greece} \\
  \texttt{\{adamsp,arly\}@uoi.gr}
}

\begin{document}

\maketitle

\begin{abstract}
Neural classifiers are non linear systems providing decisions on the classes of patterns, for a given problem they have learned. The output computed by a classifier for each pattern constitutes an approximation of the output of some unknown function, mapping pattern data to their respective classes. The lack of knowledge of such a function along with the complexity of neural classifiers, especially when these are deep learning architectures, do not permit to obtain information on how specific predictions have been made. Hence, these powerful learning systems are considered as black boxes and in critical applications their use tends to be considered inappropriate. Gaining insight on such a black box operation constitutes a one way approach in interpreting operation of neural classifiers and assessing the validity of their decisions. In this paper we tackle this problem introducing a novel methodology for discovering which features are considered relevant by a trained neural classifier and how they affect the classifier's output, thus obtaining an explanation on its decision. Although, feature relevance has received much attention in the machine learning literature here we reconsider it in terms of nonlinear parameter estimation targeted by a set membership approach which is based on interval analysis. Hence, the proposed methodology builds on sound mathematical approaches and the results obtained constitute a reliable estimation of the classifier's decision premisses.     
\end{abstract}

\section{Introduction}
\label{section:1}

Performance success of machine learning (ML) systems is mainly due to nonlinear models used for modeling the processes governing real world problems. These models result from complex nonlinear ML approaches such as neural networks, kernel-based techniques, boosting, Gaussian processes etc. Among these approaches the most recent paradigm concerns Deep Learning (DL), a class of neural network architectures consisting of several interconnected layers with numerous neurons. These stacked architectures are designed to carry out a number of nonlinear operations in order to perform complex pattern recognition operations on large data sets such as images and video or data related to human action recognition, natural language processing etc.

However, most of the above efficacious ML algorithms have a major drawback of lacking transparency in their decision mechanism. It is common knowledge that nonlinear classifiers and in particular neural classifiers act as black boxes as they provide no means of direct human understanding of their reasoning. This constitutes a serious defect as humans using these systems cannot rely upon them without understanding the learned elements of their reasoning in order to interpret and verify decisions. Moreover, this proves to be a problem especially in cases where safety and operation of critical systems is at stake or when explicit legal requirements are dictated, such as the General Data Protection Regulation (GDPR) recently adopted  by the the European Union. This renewed GDPR imposes legal requirements on the use of Artificial Intelligence (AI) systems either for profiling or for automated decision-making purposes. Among these requirements {\em{Transparency towards individuals}} stands for providing meaningful information explaining the results of an AI system processing personal data \cite{GDPR2017}.

Interpreting intelligent operation has been a matter of concern for the AI community since the era of symbolic knowledge processing \cite{Moore1988}. Earlier approaches to explainability of AI systems include rule extraction techniques \cite{ANDREWS1995373}, inductive logic programming \cite{Muggleton1995}, association rule learning \cite{agrawal1993mining} etc. Linear systems while not suitable for dealing with complex learning tasks proved to be sufficiently explicative of their operation offering a basis for triggering research in explainable-interpretable ML systems. Though interpretability of learned systems proved to be useful in several scientific and technological areas \cite{bayani2021fanoos,Clarke2003VerificationOH,Wen1996,Kearfott1996} it was the advent of DL that, actually, established the topic known as eXplainable Artificial Intelligence (XAI). 

\subsection{Related Work}
\label{subsection:1.1}

In the early times of symbolic knowledge based systems such as the Expert Systems the means for providing explanations about the specific line of thought leading to conclusions was based on reporting the detailed triggering and chaining of rules from their premisses to the deductions. When neural networks became popular, as tools for intelligent tasks, work on their explainability-interpretability highlighted the problem of dealing with the black box operation of these systems. So, earlier research efforts on this topic tried to tackle explainability in terms of extracting rules from neural network operation \cite{ANDREWS1995373}.  
   
Recent research on XAI has received increased attention due to initiatives such as the DARPA XAI project and the Explainable Artificial Intelligence Workshops held during the IJCAI-2017 \cite{IJCAI17} and IJCAI-2019 \cite{IJCAI19}. Moreover, it is important to mention the statement on algorithmic transparency and accountability issued by ACM which prompts for principles that should be applied to algorithms for automatic decision-making. A number of surveys have been already published covering relevant research on XAI, \cite{Samek2019}, \cite {BARREDOARRIETA202082}, \cite{guidotti2018survey}. Among these, here, we retain the taxonomy and terminology of the survey paper of \cite{Burkart2021ASO} which elaborates on several aspects of the explainability matters while referring to the most significant research and related results. 

Classifiers and learners operating with interpretable models, such as decision trees, rule-based or linear models, Bayesian models and k-nearest neighbors can be considered as interpretable by nature. On the other hand classifiers disposing a ``white box'' structured model can be trained in order to provide explanations and so they are explainable by design. These models include decision trees, decision rules, Bayesian list machines etc. For all these the interested reader may refer to \cite[section~{4}]{Burkart2021ASO} and references therein.  

Here, we are mostly concerned with research on classifiers characterized by black box operation. Several different approaches have been proposed for tackling the explanation-interpretation of the decisions learned by a learner \cite{guidotti2018survey, henelius2014peek}. With respect to the analysis given in \cite[section~{5}]{Burkart2021ASO} we note the following; some approaches cope with the model itself, either trying to explain the model as a whole or by replacing it with another model (surrogate) which is inherently understandable. For classification tasks one may cite the work of \cite{Ribeiro2016ModelAgnosticIO, ribeiro2016should} who proposed a linear model called SP-Lime, a tree-based approach  \cite{Frosst2017DistillingAN} called Soft Decision Tree, as well as various {\em{tree-based}} approaches such as \cite{schetinin2007confident, hara2016making, yang2018global}. Another important class of surrogate model approaches are based on rule extraction. These rule models adopt various extraction methods and their scope is either pedagogical or decompositional with the latter been largely decision rules from Support Vector Machines and Artificial Neural Networks, see \cite{martens2011performance}, \cite[section~{5}]{Burkart2021ASO} and references therein. While the previous approaches are based on surrogate models acting at a global level for the classifier other approaches are based on a so-called local surrogate model which provides valid explanations for specific data instances and their vicinity, such as LIME \cite{ribeiro2016should} and anchorLIME \cite{ribeiro2018anchors}, Model Understanding through Subspace Explanations (MUSE) \cite{lakkaraju2019faithful}, SHAP (Kernel SHAP or Linear SHAP)\cite{lundberg2017unified}, etc.

Contrary to the above approaches a family of methods proceed to generating explanations directly from the black box model without resorting to an intermediary surrogate model. These methods are further subdivided to those adopting a global explanation generation and those who generate local explanations. While the former tend to provide explanations regardless of the specific model predictions and they try to reveal significant properties of the black box model the latter act at the level of specific predictions of the learner and their explanations are valid in the vicinity of some prediction. Methods relying on feature importance are the most relevant among global explanation generation methods. Among them one may cite, Random Forest Feature Importance and Leave-One-Covariate-Out (LOCO) Feature Importance \cite{hastie2009elements, hall2017machine}, Quasi Regression \cite{jiang2002quasi}, Global Sensitivity Analysis (GSA) \cite{cortez2011opening}, Gradient Feature Auditing (GFA) approach \cite{adler2018auditing}, Partial Dependence Plots (PDP) \cite{goldstein2015peeking} etc. 

Local explanation methods are grouped into saliency methods, counterfactual ones, inverse classification and prototypes and criticism. Saliency methods generate explanations of specific model predictions by taking into account the explanatory power of the feature vector, i.e. the salience of the individual features. According to \cite{Burkart2021ASO} two main trends have been formed in this group; one dealing with feature attribution that is methods ``which explicitly try to quantify the contribution of each individual feature to the prediction'' while the second builds on attention based models which given a prediction of a classification task try to reveal the most relevant parts of input features which support this prediction; \cite{bahdanau2014neural} propose the formation of of context vectors, while the Attention Based Summarization (ABS) system ``summarizes a certain input text and
captures the meaningful paragraphs'' \cite{rush2017neural}.

Among local explanation methods based on feature attribution we may cite the following: Local Gradients approach \cite{baehrens2010explain} which learns a local explanation vector consisting of local gradients of the probability function, Quantitative Input Influence (QII) \cite{fisher1801model}, Shapley values \cite{shapley1951notes}, Self Explaining Neural Networks (SENN) \cite{alvarez2018towards}. DeepLIFT (Deep Learning Important FeaTures) \cite{shrikumar2017learning} uses back-propagation of a prediction to the feature space in order to generate a reference
activation. Deep Taylor proposed by \cite{montavon2017explaining} applies to Deep Learning architectures visualizing in heat maps which features contribute to a specific output, Layerwise Relevance Propagation \cite{samek2017explainable} uses the prediction of a model and applies redistribution rules to assign a relevance score to each feature. In \cite{samek2017explainable} the authors introduce the Sensitivity Analysis (SA) and Layer-wise Relevance Propagation (LRP) approaches for providing explanation on the predictions of black box models in terms of input variables. In the SA approach the output of a model is explained using the model's gradients and the approach constructs a heat map visualizing the effect of features' values on the classification score. 

While there are several interesting approaches based on feature attribution here we retained the most significant between them in terms of similarity with the proposed approach. Actually, the approach elaborated in this paper is  mainly a local explanation method which reveals the contribution of features to specific data instances. However, as we will show, due to its set oriented character, the approach may provide a broad view of the feature attribution referring to more data instances as well as to significant areas of the pattern space. A significant advantage of the approach is that it relies exclusively on operation between sets and does not make use of gradients. The rest of the paper is organized as follows. In Section 2, we present the concepts on which the proposed approach is built. Section 3 is devoted to a presentation of the approach itself with details on its application. In Section 4 we perform an experimental study using well known benchmarks and we discuss advantages and pitfalls detected in this work. Finally, Section 5 closes the paper with some comments and concluding remarks.

\section{The Proposed Approach}
\label{section:2}

\subsection{Set Membership Estimation}
\label{subsection:2:1}

This is a simplified presentation of the approach given in \cite{milanese1991optimal} and in particular as far as the solution operator is concerned which in our presentation is the identity operator $S(\cdot)=I$. Typically, for an estimation problem one wishes to evaluate some unknown variables based on measurements performed on some real process. These measurements are always contaminated with noise inducing uncertainty on the estimated variables. Here, the estimated variables are considered to be elements of a set $X$, called the problem element space, the measurements are elements of a set $Y$ and there exists a mapping ${\mathcal{F}}:X \rightarrow Y$, called the information operator, which for any element of $X$ returns all the measurements corresponding to this element. These sets will be considered metric linear spaces with dimensions $n$ and $m$ and respective norms ${\| \cdot \|}_{X}$ and ${\Vert \cdot \Vert}_{Y}$. Given a set $K \subset X$ and ${\mathbf{x}} \in K$ the information operator issues ${\mathcal{F}}({\mathbf{x}})$, a quantity which is meant to be measured in $Y$ and, usually, is not exactly known, since it is the outcome of a measurement process corrupted by some noise ${\mathbf{e}}$ which constitutes the error of the measurement. Then, assuming that noise is additive the measurement ${\mathbf{y}}$ is given by
\begin{equation}
	{\mathbf{y}} = {\mathcal{F}}({\mathbf{x}})+{\mathbf{e}}.
\end{equation}

Even in the absence of any other information on the noise (error) model, such as its distribution or other statistical properties, an important requirement set for the additive noise is that it is bounded in $Y$, i.e. ${\Vert {\mathbf{e}} \Vert}_{Y} \leqslant \varepsilon$ for some known value of $\varepsilon$. Then the estimation problem is considered to be a bounded-error estimation problem. 

The estimation problem consists in determining an estimation algorithm $\Phi$ (an estimator) that is a mapping $\Phi:Y \rightarrow X$ which given some measurement ${\mathbf{y}} \in Y$ provides an approximation $\Phi({\mathbf{y}}) {\triangleq} {{\mathcal{F}}^{-1}({\mathbf{y}})} \approx {\mathbf{x}}$. Here, despite the notation used, we need to note that, in general, ${\mathcal{F}}$ is not invertible in the classical sense. When $\Phi(\cdot)$ provides as output a single value in $K$ then it is a {\em{point estimator}} otherwise it is a {\em{set estimator}} as the estimation yields a set $\Phi({\mathbf{y}}) \subset K$. Work presented in this paper deals with set membership estimation. 

In the absence of relevant information about the input value ${\mathbf{x}}$ producing ${\mathbf{y}}={\mathcal{F}}({\mathbf{x}})$ a direct consequence of the bounded error ${\mathbf{e}}$ hypothesis is that there exist, eventually, multiple inputs, i.e. a set of, ${\mathbf{x}} \in K$ giving ${\mathcal{F}}({\mathbf{x}})$ at a distance from ${\mathbf{y}}$ less than or equal to $\varepsilon$. Hence, for devising an efficient set estimation algorithm one needs to define and effectively characterize the sets involved in the estimation process both in the measurement space $Y$ and in the problem input space $X$ with respect to the error ${\mathbf{e}}$. So, given a set $K \subset X$ of feasible input values, a model related information function ${\mathcal{F}}$, a measurement ${\mathbf{y}} \in Y$ and the error ${\mathbf{e}}$ in measurements we may consider the following sets:
\begin{itemize}
	\item {$M_{{\mathbf{y}}} = \{\hat{{\mathbf{y}}} \in Y : {\Vert {\mathbf{y}}-\hat{{\mathbf{y}}} \Vert}_{Y} \leqslant \varepsilon\} $. \\ The set of measurements at a distance bounded by $\varepsilon$ from a measurement ${\mathbf{y}}$.}
	\item {$F_{{\mathbf{y}}} = \{{\mathbf{x}} \in K : {\Vert {\mathbf{y}}-{\mathcal{F}}({\mathbf{x}}) \Vert}_{Y} \leqslant \varepsilon\} $. \\ The set of feasible input values related to a measurement ${\mathbf{y}}$ and the error bound about it.}
	\item {$E_{\Phi,{\mathbf{y}}} = \{{\mathbf{x}} \in K : {\mathbf{x}} \in {\Phi({\mathbf{y}})} \} $. \\ The set of estimated inputs provided by the estimation algorithm for some measurement and the error bound about it. }
\end{itemize}

It is obvious that the objective of the estimation algorithm $\Phi(\cdot)$ is to derive $E_{\Phi,{\mathbf{y}}}$, for ${\mathbf{y}} \in Y$. Whenever the estimation algorithm is such that $E_{\Phi,{\mathbf{y}}}=F_{{\mathbf{y}}}$ we have an exact estimation algorithm while otherwise $E_{\Phi,{\mathbf{y}}}$ or, equivalently, $\Phi({\mathbf{y}})$ will be the result of an approximation procedure. Note that formation of the above sets depends, mainly, on the model related information function ${\mathcal{F}}$ which may result in complex sets such as non-connected and/or non-convex. So, when designing a set estimation algorithm to approximate $E_{\Phi,{\mathbf{y}}}$ one needs to consider this remark. In general, approximation of $E_{\Phi,{\mathbf{y}}}$ can be done by constructing sets which constitute either inner or outer approximations. Typical application of the above set membership approach concerns estimation of model parameters from measurements under the assumption that uncertainty in parameters is bounded. This is what, later in this section, is referred as parameter estimation problem in a bounded-error context. 


Techniques proposed in the literature to compute the approximated input area, comprise the use of different solids such as ellipsoids, orthotopes, parallelotopes etc depending on the dimension of the problem and the norms chosen for $Y$ and $X$. The approach used throughout this paper, to devise an estimation algorithm $\Phi$ and to estimate the area corresponding to $\Phi({\mathbf{y}})$, builds on multidimensional intervals that is, axis-aligned boxes (hyper-boxes). There are at least two reasons for adopting these constructs. The first deals with the ability of intervals to quantify bounded uncertainty, i.e. error, about parameters of the model function without specific assumptions on their distribution or any other restriction. The second has to do with the solid mathematical background on which intervals are founded, that is Interval Analysis (IA),  which permits to consider intervals in several dimensions and functions on these interval constructs transcribing and applying numerous results and algorithms from numerical analysis.   

\subsection{Interval Analysis}
\label{subsection:2:2}

An interval or interval number is a closed interval $I=\left[a,b\right] \subset {\mathbb{R}}$ of all real numbers between (and including) the endpoints $a$ and $b$, with $a \leqslant b$. Note that if $a=b$ then the interval is a point, or else degenerate or thin interval. Usually, a real interval is denoted $[\underline{x},\overline{x}]$ or $[x]$. For a real interval $[\underline{x},\overline{x}]$ the condition $\underline{x}=-\overline{x}$ implies that this is a symmetric interval. The arithmetic defined on intervals is called interval arithmetic. Interval real numbers are elements of ${\mathbb{IR}}$ and can be used to compose interval objects such as vectors and matrices. So, a $n$-dimensional interval vector $\mathbf{V}$ is a vector having $n$ components $(\mathbf{v_1},\mathbf{v_2},\ldots,\mathbf{v_n})$ such that every component $\mathbf{v_i}, 1 \leqslant i \leqslant n$ is a real interval $[\underline{v_i},\overline{v_i}]$. The set on $n$-dimensional vectors of real intervals is denoted ${\rm {\mathbb I}{\mathbb R}^{n}}$. Interval matrices are defined in the same way and finally interval functions can be defined for describing mappings between interval objects as it happens with their real counterparts. The subsequent study of these interval objects and functions resulted in the establishment of IA \cite{alefeld2000interval}. 

Interval arithmetic was introduced as a means to perform numerical computations with guaranteed accuracy and bounding the ranges of the quantities, used in computations. From another point of view, when measuring various quantities or estimating parameters, usually one needs to cope with errors due to imperfect-defective measurements or uncertainty in estimation. Hence, enclosing real values between bounds is equivalent to quantifying uncertainty and to bounding errors on these values. IA methods permit to control error propagation through computations and thereby uncertainty about parameters diffused in computations so that the algorithms provide results verified by design.   

In practical calculations interval arithmetic operations are reduced to operations between real numbers \cite{JaulinBook}. For the intervals ${[\underline{x},\overline{x}],[\underline{y},\overline{y}]}$ it can be shown that the following intervals are produced for each arithmetic operation:

\begin{subequations}
\begin{align} 
[\underline{x},\overline{x}] + [\underline{y},\overline{y}] &= [\underline{x}+\underline{y},\overline{x}+\overline{y}]  \\ 
[\underline{x},\overline{x}] - [\underline{y},\overline{y}] &= [\underline{x}-\overline{y},\overline{x}-\underline{y}]  \\ 
[\underline{x},\overline{x}]\times[\underline{y},\overline{y}] &= \left[\min \left(\underline{x}\underline{y},\underline{x}\overline{y},\overline{x}\underline{y},\overline{x}\overline{y}\right),\max \left(\underline{x}\underline{y},\underline{x}\overline{y},\overline{x}\underline{y},\overline{x}\overline{y}\right)\right]  \\ 
[\underline{x},\overline{x}]\div{[\underline{y},\overline{y}]} &= [\underline{x},\overline{x}]\times \dfrac{1}{[\underline{y},\overline{y}]},\,\rm{with}  \\
\dfrac{1}{[\underline{y},\overline{y}]} &= \left[\dfrac{1}{\overline{y}},\dfrac{1}{\underline{y}} \right],\,{\rm{provided\,that\,}\, 0} \notin \left[\underline{y},\overline{y}\right]  
\end{align}
\end{subequations}

Given a real function $f:D \subset \mathbb{R} \rightarrow \mathbb{R}$ and an interval $[x] \subseteq D$ in its domain the range of values of $f$ over $[x]$ is denoted by $f([x])$. Computation of $f([x])$ using IA tools consists in enclosing it by an interval which is as narrow as possible. This constitutes an important matter in IA as various problems can be formulated in these terms: localization and enclosure of global minimizers of $f$ on $[x]$, verification of $f([x]) \subseteq [y]$ for given $[y]$, nonexistence of a zero of $f$ in $[x]$ etc.  

\begin{figure*}[t!]
	\centering
	\includegraphics[width=3.5in]{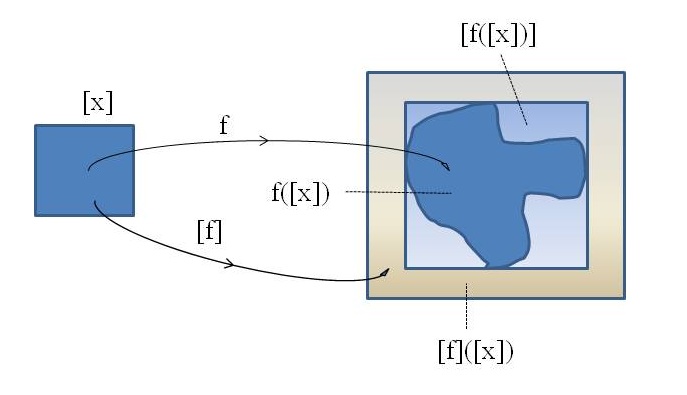}%
	\caption{\small{A function $f$, an inclusion function $[f]$ and the images of $[x]$}}
	\label{fig1.0}
\end{figure*}

Enclosing $f([x])$ requires to define a suitable interval function $[f]:{\rm {\mathbb I}{\mathbb R}} \rightarrow {\rm {\mathbb I}{\mathbb R}}$ such that $\forall [x] \in {\rm {\mathbb I}{\mathbb R}},\,\,\,f([x]) \subset [f]([x])$, see Figure \ref{fig1.0}. This interval function $[f]$ is called an inclusion function of $f$ and makes it possible to compute a box $[f]([x])$ which is guaranteed to contain $f([x])$, whatever the shape of $f([x])$. If $f(x), x \in D$ is computed using a finite composition of elementary arithmetic operators $\{+,-,\times,\div\}$ and standard functions such as \{exp,sqr,cos,sin,..\}, then, the inclusion function which is obtained by replacing in $f$ the real variable $x$ by an interval variable $[x] \subseteq D$ and each operator or standard function by its interval counterpart, is called a \textit{natural inclusion function} of $f$. As noted in \cite{JaulinBook} the natural inclusion function has important properties such as being \textit{inclusion monotonic}, and if $f$ is defined using only continuous operators and continuous standard functions the natural inclusion function is convergent. The above can be generalized to functions $f: {{\mathbb{R}}^{n}} \rightarrow {{\mathbb{R}}^{m}}$.
 
\subsection{Set Approximation Using Interval Analysis}
\label{subsection:2:3}

Set membership approximation using IA was investigated by \cite{JaulinWalter93} who introduced the method known as Set Inversion Via Interval Analysis (SIVIA). The method was originally proposed in order to allow for the guaranteed estimation of nonlinear parameters from bounded error data and has been applied to numerous approximation problems described either by static or dynamic models. The method results in defining unions of axis-aligned boxes approximating a set of interest. As previously, given a function $f:X \rightarrow Y$, where $X \subset {\rm \mathbb{R}}^{n}$, $Y \subset {\rm \mathbb{R}}^{m}$ and an interval vector, i.e. a box, $[{\mathbf{y}}] \subseteq Y$, the objective is to define the set of unknown vectors ${\mathbf{x}} \in X$ such that $f({\mathbf{x}}) \in [{\mathbf{y}}]$. This set can be defined as $K = \{{\mathbf{x}} \in X \subseteq {\rm \mathbb{R}}^{n}\,|\,f({\mathbf{x}}) \in [{\mathbf{y}}]\} = f^{-1}([{\mathbf{y}}]) \cap X$, where $X$ is the search space containing the set of interest $K$; $[{\mathbf{y}}]$ is known in advance to enclose the image of the set $f(K)$ and $K$ denotes the unknown set of interest. Note that, inverse notation, $f^{-1}$ is used to denote the reciprocal image of $f$, as $f$ may not be invertible in the classical sense. 

For the set approximation problem SIVIA computes the unions of boxes denoted $K^{-}$, ${\rm{\Delta}}K$ which correspond to the inner and the boundary approximations of $K$. An additional approximation $K^{+}$ is computed by the operation $K^{+}=K^{-} \cup {{\rm{\Delta}}K}$. So, $K^{-}$, $K^{+}$ form guaranteed inner and outer enclosures of $K$ as they satisfy the relation ${K^{-}} \subseteq K \subseteq {K^{+}}$, \cite{JaulinBook}. SIVIA explores the whole search space in a branch-and-bound strategy. During computation, a box $[{\mathbf{x}}] \in X$ receives one of three possible designations, 
\begin{enumerate}[label=\Alph*]
	\item {feasible if $[{\mathbf{x}}] \subseteq K$ and $[f]([{\mathbf{x}}]) \subseteq [{\mathbf{y}}]$}, 
	\item {infeasible if $[f]([{\mathbf{x}}]) \cap [{\mathbf{y}}] = \emptyset$ and}, 
	\item {indeterminate if $[{\mathbf{x}}]$ may be feasible, unfeasible or ambiguous}. 
\end{enumerate}
The condition $[{\mathbf{x}}] \subseteq K$ and $[f]([{\mathbf{x}}]) \subseteq [{\mathbf{y}}]$ is necessary and sufficient for $[{\mathbf{x}}]$ to be feasible. Feasible boxes are added to $K^{-}$ and infeasible become members of the complement of $K^{+}$ denoted $N$. Finally, any indeterminate box is bisected and the method recursively examines the two resulting sub-boxes. Bisection is possible up to some limit, which is preset for the problem and defines its resolution. Boxes that are indeterminate and cannot be further bisected are added to ${\rm{\Delta}}K$. For a detailed description of the algorithm implementing SIVIA the reader may refer to \cite{JaulinBook}. It is worth noting here that SIVIA applies to any function $f$ for which an inclusion function $[f]$ can be computed. The algorithm \ref{InitAlgo} below provides a formal description of the method. 

\renewcommand{\algorithmiccomment}[1]{// #1}
\label{sivia_algorithm}
\begin{algorithm}[!ht]
\caption{SIVIA(in: $[f]$, $[{\mathbf{y}}]$, $\varepsilon$, $[{\mathbf{x}}]$ ; inout: $K^{-}$, ${\rm{\Delta}}K$, $K^{+}$, $N$)}
\label{InitAlgo}
\begin{algorithmic}[1]
\REQUIRE {Before calling SIVIA initialize $[f]$, $[{\mathbf{y}}]$, $\varepsilon$; \\ 
and set ${K^{-} := \emptyset}$; ${{\rm{\Delta}}K := \emptyset}$}; ${K^{+} := \emptyset}$, ${N := \emptyset}$;
\IF {$[f]([{\mathbf{x}}]) \subset [{\mathbf{y}}]$}
  \STATE {${K^{-}}:={K^{-}} \cup [{\mathbf{x}}]$;}
	\STATE {${K^{+}}:={K^{+}} \cup [{\mathbf{x}}]$;}
	\STATE {return;}
\ENDIF
\IF {$[f]([{\mathbf{x}}]) \cap [{\mathbf{y}}] = \emptyset$}
	\STATE {${N} := {N} \cup [{\mathbf{x}}]$;}
	\STATE {return;}
\ENDIF
\IF {${\rm{width}}([{\mathbf{x}}]) < \varepsilon$}
	\STATE {${\rm{\Delta}}K:={\rm{\Delta}}K \cup [{\mathbf{x}}]$;}
	\STATE {${K^{+}}:={K^{+}} \cup [{\mathbf{x}}]$;}
	\STATE {return;}
\ENDIF
\STATE {bisect $[{\mathbf{x}}]$ getting left and right sub-boxes $[{\mathbf{x}}]_L$ and $[{\mathbf{x}}]_R$} 
\STATE SIVIA($[f]$, $[{\mathbf{y}}]$, $\varepsilon$, $[{\mathbf{x}}]_L$, ${K^{-}}$, ${\rm{\Delta}}K$, ${K^{+}}$, $N$);
\STATE SIVIA($[f]$, $[{\mathbf{y}}]$, $\varepsilon$, $[{\mathbf{x}}]_R$, ${K^{-}}$, ${\rm{\Delta}}K$, ${K^{+}}$, $N$);
\end{algorithmic}
\end{algorithm}

\section{Estimating Feature Relevance of Neural Classifiers}
\label{section:3}

\subsection{Formulation of the Set Membership Problem}
\label{subsection:3:1}

Set membership estimation and more specifically SIVIA is, typically, used in model and parameter identification problems of either static or dynamical systems. The approach was effectively applied in two specific problems concerning neural classifiers; first \cite{adam2015reliable} used SIVIA in order to estimate the domain of validity of a Multilayer Perceptron (MLP) trained on some classification problem. In this paper the MLP function is considered to be a non-linear static model of the problem's classification mapping and the input space is dimensioned by the features. The results obtained not only demonstrated the application of set membership estimation on a neural classifier such as a trained MLP but also proved that the resulting estimation is verified by design because of the use of IA for the computation of the estimated sets. Based on the reliable estimation of the domain of validity of a trained MLP \cite{adam2019evaluating} showed that it is possible to derive measures of its generalization performance in a deterministic way without the uncertainty induced by statistical estimations such as cross validation. 

However, while the previous works showed that it is possible to reliably estimate the dependence of the output prediction of a trained MLP on the values of the input patterns, these successful applications entail a significant amount of computational cost related to the exhaustive investigation of the input space by the branch-and-bound  strategy of SIVIA. This computational cost becomes intolerable for problems with medium to large size input dimensions, i.e. 10 parameters. On the other hand, retaining the reliable estimation of the feasible input set which steers the classifier's output while overcoming the computational burden is a goal that can be achieved by adequately reducing the size of the search space. This is done by considering some specific input dimensions and evaluating the contribution of the respective features to the output produced by the classifier for any given pattern.

To be more specific, let us suppose that the classification function implemented by the neural classifier after training is denoted by ${\mathcal{F}}:X \rightarrow Y$, where $X$ is the input data space, consistently represented by the sample pattern data and $Y$ is the set of discrete labels associated with the classes of the problem. Moreover, without loss of generality, let us consider that a $1$-of-$M$ scheme is used for coding the response of the classifier for the associated classes. Then, it is well known that, except for perfect training, the output of neural classifiers with sigmoid or softmax output nodes, very often, does not give the exact values expected for the pattern classes. So, when some input pattern ${\mathbf{x}}$ with class label $j$ is correctly classified the $j$-th output node gives a prediction in some interval $[1-{\beta}_{j},1]$ which is a compact set denoted $O_{j}({\mathbf{x}})$. Given these sets, for all patterns, it is possible to consider an axis aligned box, $[Y]$, which contains all these sets and so it encloses the output space $Y$. 

Furthermore, let us suppose that $[{\mathcal{F}}]$ denotes the interval inclusion function defined for ${\mathcal{F}}$ and $[X]$ is an axes aligned box which encloses the pattern space $X$. Unless otherwise defined, $[X]$ is the interval vector $[X]=([X_{1}],[X_{2}],\ldots,[X_{n}])$. Then, we may write
\begin{equation} 
	[{\mathcal{F}}]:[X]\rightarrow [Y].
\end{equation}
  
When some specific prediction of the classifier is given and the area of the input space triggering this prediction is required, the above formulation clearly defines a set membership estimation problem that can be tackled using SIVIA. 

In this context explaining the prediction $j$ provided by the classifier for the input pattern ${\mathbf{x}}=(x_{1},x_{2},\ldots,x_{n}) \in X$ requires to consider the interval vector $[{\mathbf{x}}]=([x_{1}],[x_{2}],\ldots,[x_{n}]) \subseteq [X]$ enclosing the values of the features and to estimate the set of values contributing to the prediction. This interval vector forms the set of feasible input values, or else, following notation of subsection \ref{subsection:2:1} above, it is related to the error about the output set $O_{j}({\mathbf{x}})$, i.e. the measurement. Then, in order to define the set of the input space contributing to the prediction $O_{j}({\mathbf{x}})$ we may formulate the set membership estimation problem that needs to be solved as:
\begin{equation}
	{\mathcal{F}}^{-1}(O_{j}({\mathbf{x}})) \subseteq [{\mathbf{x}}].
\end{equation}
  
This estimation problem was tackled in \cite{adam2015reliable} with all the computational burden as noted above. Herein, in order to define the contribution of some specific feature, say $k$, where $1 \leqslant k \leqslant n$, to the prediction represented by $O_{j}({\mathbf{x}})$ the set membership estimation problem that needs to be solved can be formulated as,
\begin{equation}
	{\mathcal{F}}^{-1}(O_{j}({\mathbf{x}})) \subseteq (x_{1},x_{2},\ldots,[x_{k}],\ldots,x_{n}).
\end{equation}

Note that in this formulation all intervals $[x_{i}]$, except for $i=k$, are degenerate, i.e. point intervals, corresponding to the specific values of the features for pattern ${\mathbf{x}}$. Thus, estimation of the set membership is automatically restricted to the $k$-th feature and, so, computation is absolutely tractable. 

Solving this problem results in defining the set 
\begin{equation}
{\hat{\textrm{x}}}_{k} = \{ {\hat{x}} \in [x_{k}]: {\mathcal{F}}(x_{1},x_{2},\ldots,{\hat{x}},\ldots,x_{n}) \in O_{j} \},
\end{equation}

that is, the set of values of the $k$-th feature contributing to the specific prediction. 

\begin{figure}[!ht]
	\centering
	\includegraphics[clip, trim=0.5cm 17.5cm 0.5cm 1.5cm, width=1.00\textwidth]{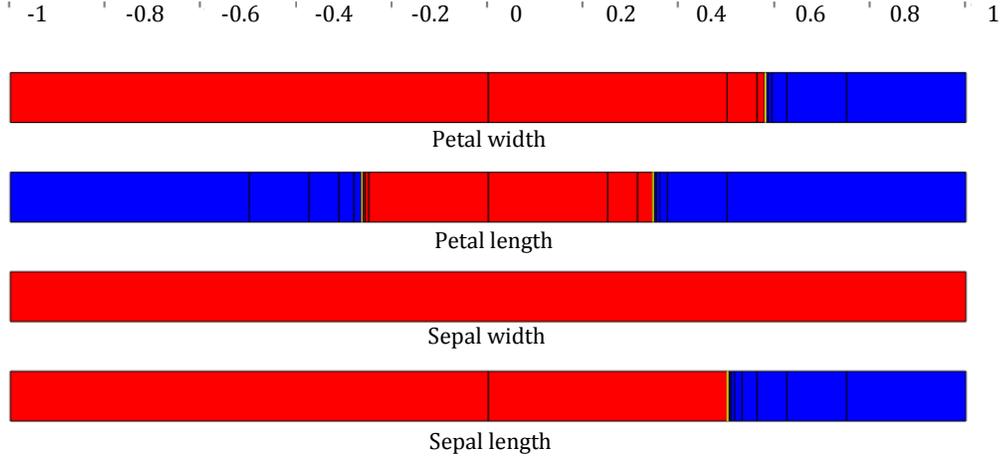}
	\caption{Examples of feature relevance derived by interval based set membership estimation.}
	\label{section3:fig1}
\end{figure}

It is important to underline, here, that it is this set ${\hat{\textrm{x}}}_{k}$ and its relation with $[x_{k}]$ that hold all the information explaining how the specific prediction $O_{j}({\mathbf{x}})$ provided by the classifier is affected by the $k$-th feature. First we need to note that ${\hat{\textrm{x}}}_{k}$ is computed as a union of subsets of $[x_{k}]$ and so ${\hat{\textrm{x}}}_{k} \subseteq [x_{k}]$. Moreover, ${\hat{\textrm{x}}}_{k}$ can be connected or not. Examples of various possible configurations of ${\hat{\textrm{x}}}_{k}$ are given in Figure \ref{section3:fig1}, hereafter. These examples concern contribution of the four features of the Fisher Iris dataset to the prediction provided by a trained MLP with $2$ hidden units for the $17$-th pattern of the class {\em{Iris-versicolor}}. 

In this problem, all features are scaled in the interval $[-1,1]$ and the interval of the classifier's output which predicts the specific class ({\em{Iris-versicolor}}) was taken to be $[0.8,1]$. Set membership estimation performed by SIVIA produced the red colored intervals for those values of the feature which are considered active as they contribute to having an output prediction in the predefined interval $[0.8,1]$ for the classifier's output dedicated to this class. On the other hand the blue areas are inactive as they do not contribute to this output. Note, that the thin yellow lines concern the ranges of values for which the contribution of the feature is undefined. These areas constitute the border between the active and non-active areas for each feature and depend on the value of the parameter $\varepsilon$ used by SIVIA, see Algorithm \ref{sivia_algorithm}, above. The thin black lines are the traces of the endpoints of the different intervals produced by the bisection applied by the SIVIA algorithm during branch-and-bound. These traces have zero length range and so they do not account for the resulting relevance. Finally, the height of the intervals displayed in this figure is added only for better visualizing the resulting intervals.  

Defining the range of the interval corresponding to the set $O_{j}({\mathbf{x}})$ for some pattern ${\mathbf{x}}$ is a determinant factor of the estimation procedure. Typically, in order to derive for some feature the range of values contributing to predict the class $j$ of the pattern one needs to consider the classifier's output, denoted $o_{j}$, define $\beta_{j}$ to be the radius of the output interval and, finally, take  $O_{j}({\mathbf{x}})=[o_{j}-\beta_{j},o_{j}+\beta_{j}]$. Obviously, when $o_{j}=1$ the interval becomes $[1-\beta_{j},1]$. The width of the interval, or else the value of $\beta_{j}$, defines the size of the set ${\mathcal{F}}^{-1}(O_{j}({\mathbf{x}}))$ or else the ``punctuality'' of the approximation of the classifier's decision and thereby to some extent the sharpness of the estimated explanations. Of course, while the size of ${\mathcal{F}}^{-1}(O_{j}({\mathbf{x}}))$ depends, mainly, on the distribution of the input data and the classifier’s model ${\mathcal{F}}$ it can be easily understood that significantly large values for $\beta_{j}$ tend to enlarge ${\mathcal{F}}^{-1}(O_{j}({\mathbf{x}}))$ towards covering the whole feasible set $X$. 

The previous considerations delineate an approach which provides significant explanations whenever the pattern is correctly classified as the prediction targeted for this pattern, i.e. $1$, is the upper bound the interval $[o_{j}-\beta_{j},o_{j}+\beta_{j}]$ or at least its distance from the upper bound is very small. On the contrary, whenever the pattern is misclassified the activation value $o_{j}$ of the respective output is a small number away from the desired interval $[1-\beta_{j},1]$. In this case, application of the set membership estimation procedure on the interval $[o_{j}-\beta_{j},o_{j}+\beta_{j}]$ will define the relevance of the features for obtaining incorrect classifications. Though this can provide the explanation to a possible question such as ``{\em{ Why the classifier provided this incorrect classification?}}'', usually, it is more useful to know ``{\em{Which are the features and consequently their values that are responsible for this incorrect classification?}}''  Then, in order to put forward an informative answer one needs to know the values of the features that contribute to a correct classification, and so, set membership estimation should compute the reciprocal image ${\mathcal{F}}^{-1}([1-\beta_{j},1])$. Here, $1$ is the desired output of the classifier while $o_{j} \neq 1$ denotes  the actual output.

\subsection{Measuring Feature Relevance}
\label{subsection:3:2}

In order to formulate a measure suitable for the feature relevance, as defined by the proposed approach, first, let us note that in qualitative terms feature relevance can be quoted as strong, weak and indifferent. As we will explain later in this subsection the term indifferent can also be interpreted as irrelevant for some feature taking part in driving a network output to be either active or inactive. With respect to this statement let us observe Figure \ref{section3:fig1} and make the following comments. 

\begin{enumerate}
	\item {It is obvious that for each feature one is able to compute the total range of the set ${\hat{\textrm{x}}}_{k}$ as the sum of the ranges of the intervals forming the union of the active part of the feature. This range is a measure $\mu$ of ${\hat{\textrm{x}}}_{k}$, denoted $\mu({\hat{\textrm{x}}}_{k})$, such that $0 \leqslant \mu({\hat{\textrm{x}}}_{k}) \leqslant \mu_{k}$, where $\mu_{k}$ denotes the length of the interval corresponding to the $k$-th feature, i.e. it denotes the value of the measure on the whole set of values for the $k$-th feature.}
	\item {A small positive value of $\mu({\hat{\textrm{x}}}_{k})$ indicates strong relevance of the feature as the total range of ${\hat{\textrm{x}}}_{k}$ is the sum of ranges of narrow intervals which correspond to strict constraints on the values of the feature. In other words the prediction is more sensitive to changes in the feature's values. This is the case with {\em{Petal length}} in Figure \ref{section3:fig1}. }
	\item {Contrary to the previous, a higher value of $\mu({\hat{\textrm{x}}}_{k})$ indicates weak relevance of the feature as there are more values of the feature which contribute to the prediction. Thus, a feature with a higher value of $\mu({\hat{\textrm{x}}}_{k})$ is less relevant to the prediction. For instance, {\em{Sepal length}} and {\em{Petal width}} are less relevant for the classifier to provide a prediction of class {\em{Iris-versicolor}} for this pattern.}
	\item {With respect to the previous comments it is obvious that the value of {\em{Sepal width}} is totally irrelevant to this specific prediction. In consequence, we may quote that, though $\mu({\hat{\textrm{x}}}_{k}) = \mu_{k}$, the relevance of the respective feature is null.}
\end{enumerate}

Following the above, it is possible to define a measure of the relevance of a feature, say the $k$-th, for a prediction provided by a neural classifier implementing a model ${\mathcal{F}}$, for any given pattern ${\mathbf{x}}$. This measure can be defined by the following formula.
\begin{equation}
\label{relevance_formula_1}
	{{\mathcal{R}}_{\mathcal{F},{\mathbf{x}},k} = 1-\dfrac{\mu({\hat{\textrm{x}}}_{k})}{\mu_{k}}}.
\end{equation}

An immediate remark that needs to be done concerns the case where $\mu({\hat{\textrm{x}}}_{k})=0$. There are two possibilities for this result. According to the first, the feature does not participate at all in the prediction and so the whole range of feature values provides output values outside the required output range which translates to a fully blue colored interval. The second possibility arises whenever a number of discrete values of the feature contribute to the required output and the algorithm used, SIVIA, is not able to detect these values due to insufficient resolution, i.e. a ``large'' value for $\varepsilon$. In this case, the algorithm stll yields a number of yellow colored intervals which indicate that there exist very small intervals of feature values which contribute to the required output. So, the decision on the feature relevance needs to take into account these yellow colored intervals.

In both cases the fraction ${\mu({\hat{\textrm{x}}}_{k})}/{\mu_{k}}$ is equal to zero and the right hand side of formula (\ref{relevance_formula_1}) yields $1$. Then, it is the total length of the undefined intervals of feature values, i.e. the yellow colored intervals, that permit to derive conclusions on the contribution of the $k$-th feature to the network output expected for pattern ${\mathbf{x}}$. Denoting ${\mu_{U}({\hat{\textrm{x}}}_{k})}$ the total length of these undefined intervals of feature values we may formulate the following two rules.
\begin{enumerate}
	\item[$\mathbf{R1}$] {If ${\mu({\hat{\textrm{x}}}_{k})}=0$ and ${\mu_{U}({\hat{\textrm{x}}}_{k})}>0$ then a number of discrete values of the feature contribute to the prediction and so ${\mathcal{R}}_{\mathcal{F},{\mathbf{x}},k} = 1$.}
	\item[$\mathbf{R2}$] {If ${\mu({\hat{\textrm{x}}}_{k})}=0$ and ${\mu_{U}({\hat{\textrm{x}}}_{k})}=0$ then the feature is totally irrelevant to this prediction and necessarily the value of ${\mathcal{R}}_{\mathcal{F},{\mathbf{x}},k}$ is taken to be equal to $0$.}
\end{enumerate} 

We need to note that, typically, we expect the case for rule $\mathbf{R1}$ to arise when a feature contributes to activating a classifier's output i.e. driving its value to an interval $[1-\beta_{j},1]$. On the contrary, rule $\mathbf{R2}$ is expected to apply when a feature is checked for its contribution to driving the classifier's output to an inactive interval of the type $[0,0+\beta_{j}]$ or $[-1,(-1)+\beta_{j}]$, depending on the activation function used for the output nodes. Then, we may conclude that a visual representation of the result of SIVIA, a sort of {\em{Relevance Map}} of the feature, permits to clarify the situation. Also, we need to note that these statements are valid for features that are not subject to be eliminated if some input pruning algorithm is applied. Experimental results, in the following section, support the validity of these statements. 

So, in order to obtain the $k$-th feature relevance the SIVIA algorithm partitions the interval $[x_{k}]$ in $N$ distinct non-overlapping intervals denoted here by, $x^{1},x^{2},\ldots,x^{N}$, each one characterized as {\em{active}} (red color), {\em{inactive}} (blue color) and {\em{undefined}} (yellow color) and thus belonging to one of three distinct families, ${\mathcal{A}}$, ${\mathcal{I}}$ and ${\mathcal{U}}$, respectively. Then, the above relation (\ref{relevance_formula_1}) as well as the two previous rules can be summarized in the following relation which can be used as a computational device for the feature relevance:

\begin{equation}
\label{relevance_formula_2}
{\mathcal{R}}_{\mathcal{F},{\mathbf{x}},k} = 
\left\{ 
	\begin{array}{lcl}
		1-\dfrac{\sum_{i = 1}^{N}{\mu(x^{i}){\mathbbold{1}_{\mathcal{A}}}(x^{i})}}{{\mu_{k}}} & \mbox{if} & {\sum_{i = 1}^{N}{\mu(x^{i}){\mathbbold{1}_{\mathcal{A}}}(x^{i})}} > 0 \\ 
		{\mathrm{H}}\left({\sum_{i = 1}^{N}{\mu(x^{i}){\mathbbold{1}_{\mathcal{U}}}(x^{i})}}\right) & \mbox{if} & {\sum_{i = 1}^{N}{\mu(x^{i}){\mathbbold{1}_{\mathcal{A}}}(x^{i})}} = 0 
	\end{array}
\right\},
\end{equation}

where $\mathbbold{1}_{\mathcal{A}}$, $\mathbbold{1}_{\mathcal{U}}$ are the indicator functions of families ${\mathcal{A}}$ and ${\mathcal{U}}$, respectively. In addition, ${\mathrm{H}}$ denotes the Heaviside function,

\[{\mathrm{H}}(z) = \left\{
  \begin{array}{lr}
    1 & \mbox{if}\quad z > 0		\\
    0 & \mbox{if}\quad z \leqslant 0 
  \end{array}
\right.
\]

Finally, it is important to note that the relevance of a feature is defined for some specific model $\mathcal{F}$ implemented by a classifier reflecting its ``reality'' concerning the classification problem.

Given the above formula the relevance computed for each feature of the specific pattern ${\mathbf{x}}$ in the example of Figure \ref{section3:fig1} is:
\begin{itemize}
	\item {Sepal length: ${\mathcal{R}}_{\mathcal{F},{\mathbf{x}},1} = 0.249451$}
	\item {Sepal width : ${\mathcal{R}}_{\mathcal{F},{\mathbf{x}},2} = 0.0$}
	\item {Petal length: ${\mathcal{R}}_{\mathcal{F},{\mathbf{x}},3} = 0.695282$}
	\item {Petal width : ${\mathcal{R}}_{\mathcal{F},{\mathbf{x}},4} = 0.209869$}
\end{itemize}

\section{Experimental Evaluation}
\label{section:4}

The experiments reported, here, were carried out on two well known benchmarks the Fisher Iris and the MNIST dataset for handwritten character recognition. For all experiments shallow neural classifiers of the MLP type were used. Details on the specific elements adopted for these neural architectures are given in the respective subsections below. The MLPs were trained using the MATLAB computing environment and for the models implemented by the trained classifiers the respective natural inclusion functions were implemented using the Ibex C++ library for interval computations, developed by Gilles Chabert et al. and released under the GNU Lesser General Public Licence (http://www.
ibex-lib.org/). SIVIA algorithm was, also, implemented using the Ibex C++ library. Finally, for the visualization of the results obtained by SIVIA, we used both the Vibes-viewer \cite{drevelle2014vibes} software library and MATLAB's plotting functions.

\subsection{Experiments on the Fisher Iris Dataset}
\label{subsection:4:1}

For these experiments the features were scaled in the interval $[-1,1]$. Besides network training purposes, this scaling was adopted in order to speed up computations performed by the SIVIA algorithm and to facilitate visualization. Two MLPs were used having a single hidden layer with 2 and 8 nodes, denoted MLP-2 and MLP-8, respectively, using the hyperbolic tangent activation function. The classification scheme adopted is $1$-of-$M$ and so there are 4 output nodes one for each class. The activation function, used for the output nodes of the MLPs, is the logistic sigmoid function. The MLPs were trained with the Levenberg-Marquardt algorithm for minimizing a Mean Squared Error function of the MLP output values with a goal set to $1e-03$. As these experiments serve to demonstrate the set membership estimation procedure the entire dataset was used for training the MLPs. For MLP-2 all patterns are correctly classified except for a single one, namely the $84$-th pattern, belonging to the {\em{Iris-versicolor}} class. An explanation of why this pattern was misclassified by MLP-2 is provided later in this subsection.

For the set membership estimation procedure the following assumptions were retained; given that predictions of MLP-2 for all patterns, except for a single one, are values very close to $1$ we defined the output interval $O_{j}({\mathbf{x}})$ for every pattern to be $[0.8,1]$. So, $\beta_{j} = 0.2$ for all $j$. Finally, the value for $\varepsilon$, the resolution parameter of SIVIA, is set to $1e-03$.

Results of the set membership estimation procedure are visualized using Relevance Maps which display for each class the contribution of all features to the prediction of the class for every pattern known to belong to this class. In the stacked presentation display of the Relevance Maps the patterns are arranged in the order they appear in the original dataset from the base up to the top of the stack, one line per pattern. Partitioning of the range of each pattern/feature in differently colored areas is done as described above \ref{subsection:3:1} with emphasis on the red colored areas which account for the values of the feature contributing to drive the output associated with the respective class to the interval $[0.8,1]$. For a direct visual estimation one may attribute a higher relevance for a pattern to the feature having a larger blue area.

\begin{figure}
\centering
\subfloat{\includegraphics[clip, trim=0.75cm 15.0cm 0.75cm 3.0cm, width=0.95\textwidth]{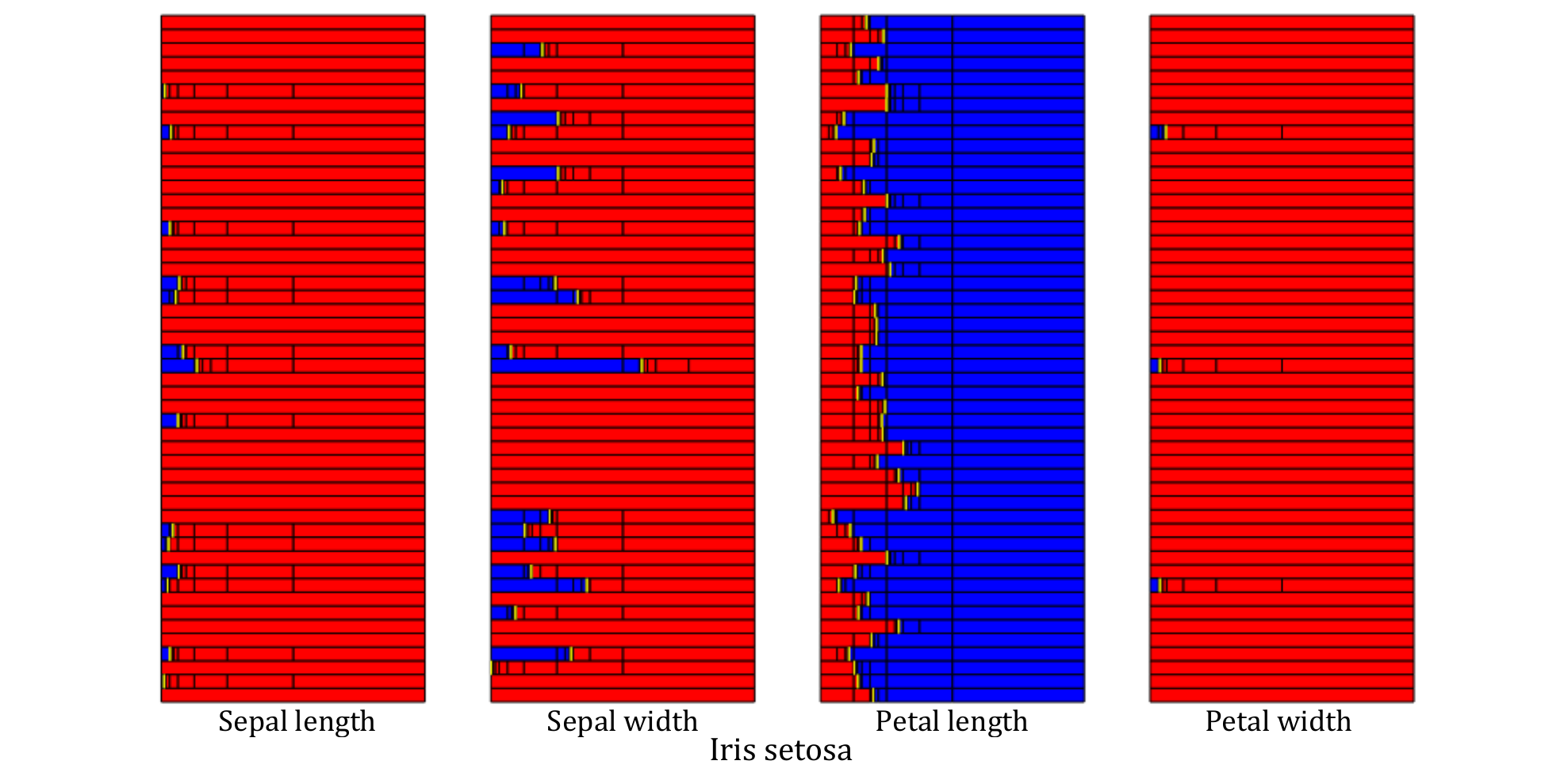}
\centering
\label{section4:fig1:1}}
\vspace{-5mm}
\vfill

\subfloat{\includegraphics[clip, trim=0.75cm 15.0cm 0.75cm 3.0cm, width=0.95\textwidth]{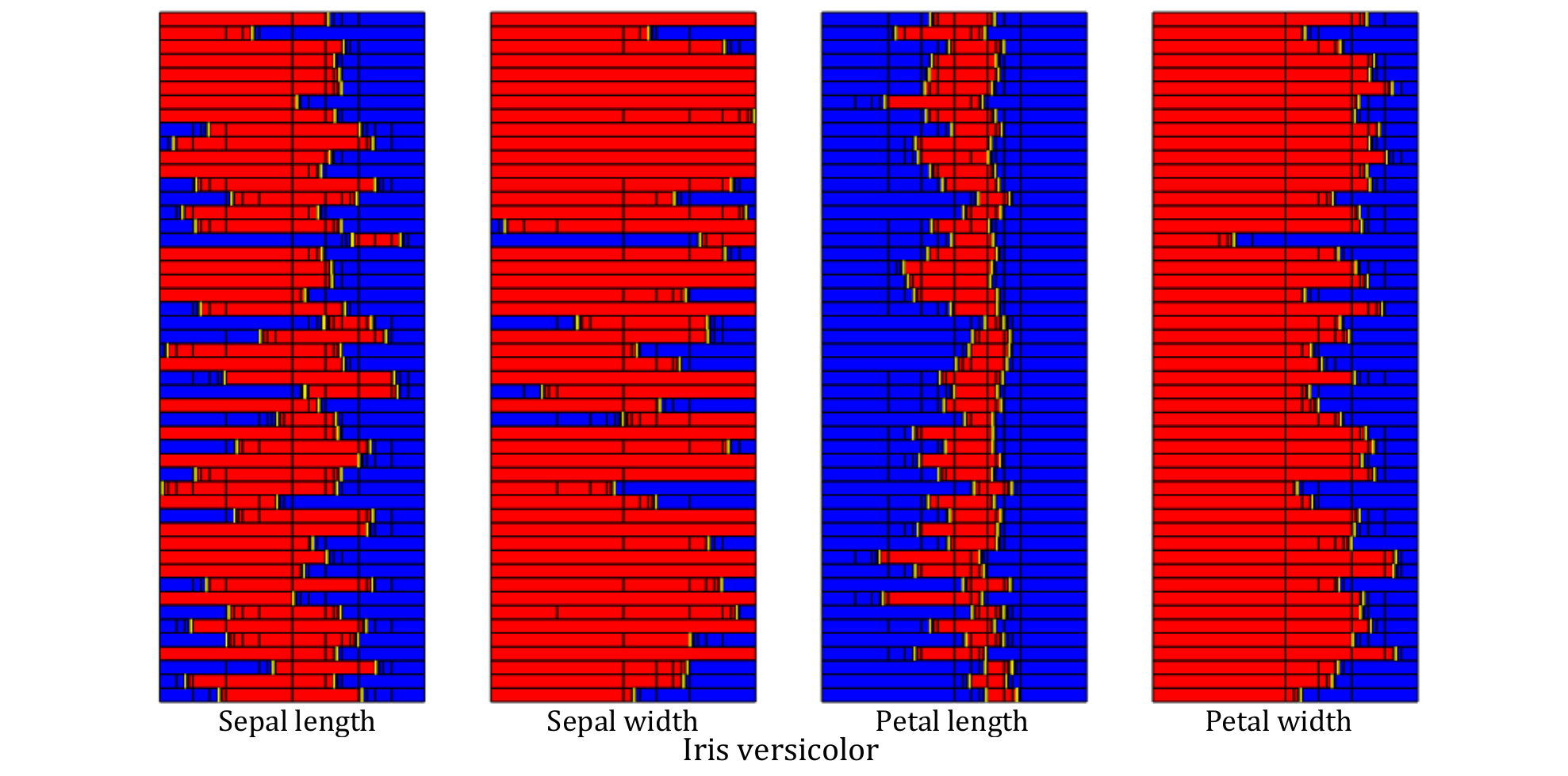}
\centering
\label{section4:fig1:2}}
\vspace{-5mm}
\vfill

\subfloat{\includegraphics[clip, trim=0.75cm 15.0cm 0.75cm 3.0cm, width=0.95\textwidth]{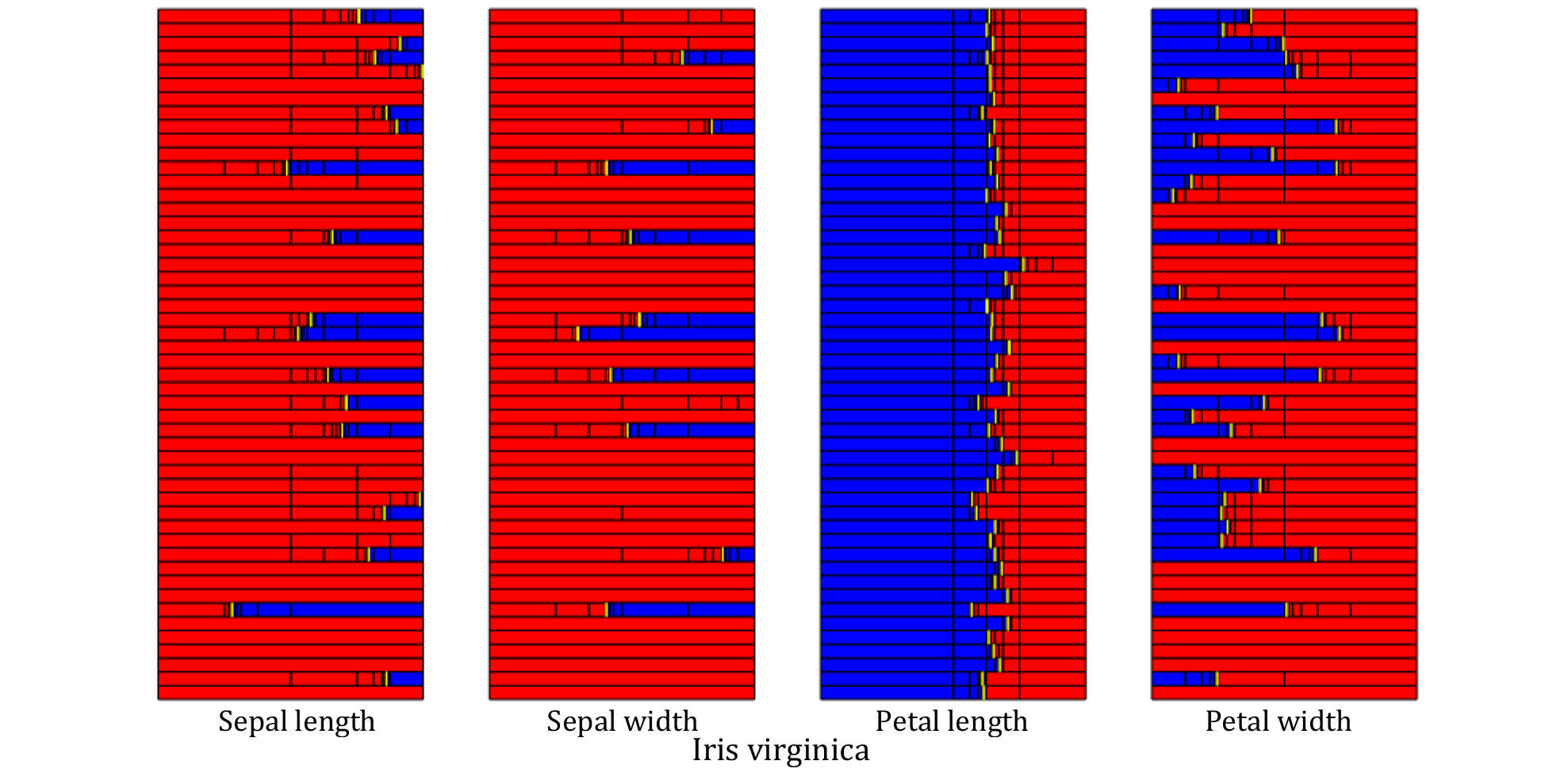}
\label{section4:fig1:3}}
\centering
\caption{{Relevance Map for the features of MLP-2. }}
\label{section4:fig1}
\vspace{-7.5 mm}
\end{figure}

Figure \ref{section4:fig1} displays the Relevance Map for all features contributing to activate the output assigned to predict the respective Iris classes for each pattern. It is worth noting that the {\em{Petal length}} is found to be the most relevant feature to predicting the patterns of all classes following the model implemented by MLP-2. 

\begin{figure}
	\centering
	\includegraphics[clip, trim=0.5cm 17.5cm 0.5cm 1.5cm, width=1.00\textwidth]{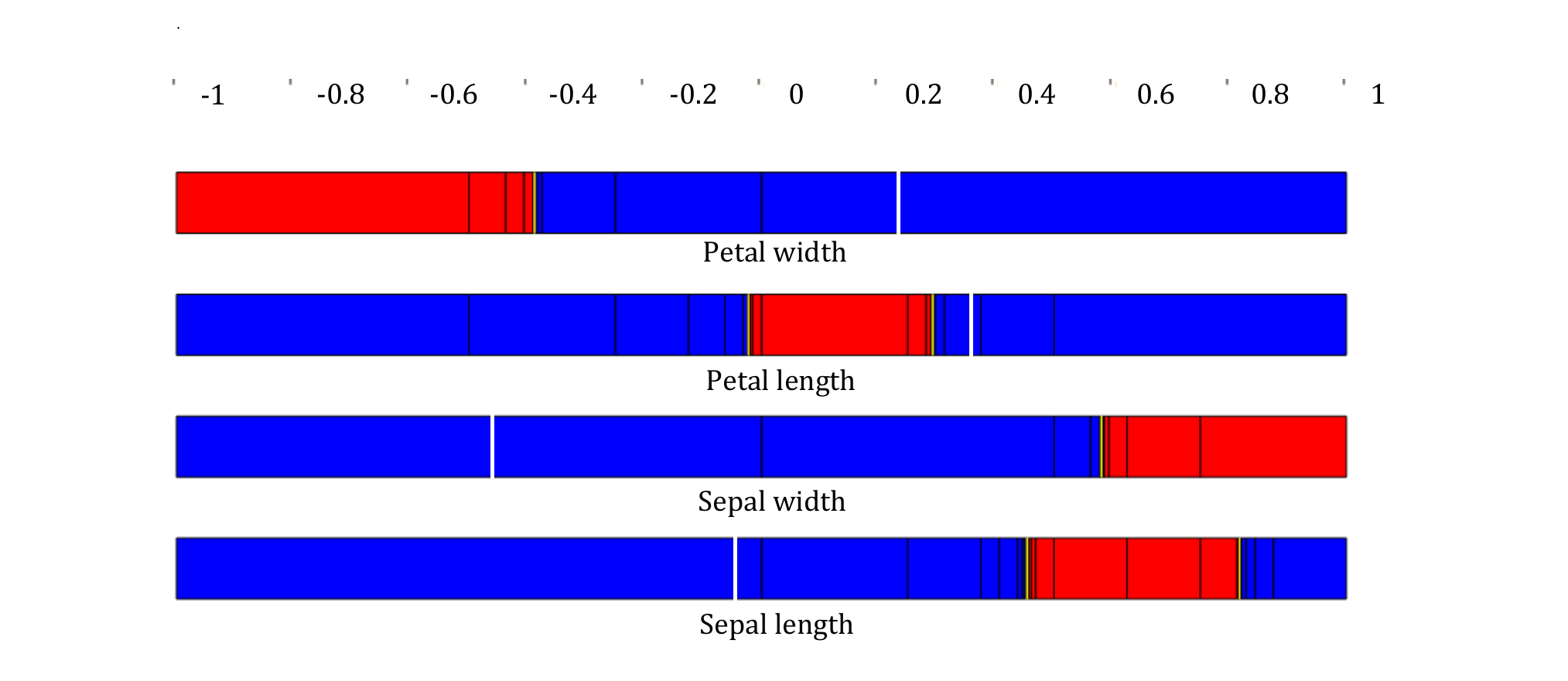}
	\caption{White traces indicate position of the feature values for a pattern in ({\em{Iris-versicolor}}) which is misclassified.}
	\label{section4:fig2}
\end{figure}

A close examination of the Relevance Map of the {\em{Iris-vesicolor}} class permits to explain why the $34$-th pattern, i.e. the $84$-th pattern of the dataset is misclassified by this classifier. Actually, the scaled values of the features for this pattern are $(-0.0556,-0.4167,0.3898,0.2500)^{T}$. Figure \ref{section4:fig2} zooms in the Relevance Map for this pattern and depicts in details the relevance of the features and the relative position of the feature values for this pattern. The red colored area for each feature is computed by SIVIA keeping the other three values fixed. The values of the features for the misclassified pattern which are noted, approximately, with white vertical traces, are clearly in the areas of the features which do not contribute to activate the output for patterns belonging to the {\em{Iris-versicolor}} class. The explanation that is possible to extract from this configuration is that the value of each feature is incompatible with the other three in terms of contributing to correctly classifying this pattern. 

\begin{figure}
	\centering
	\includegraphics[clip, trim=0.5cm 17.5cm 0.5cm 1.5cm, width=1.00\textwidth]{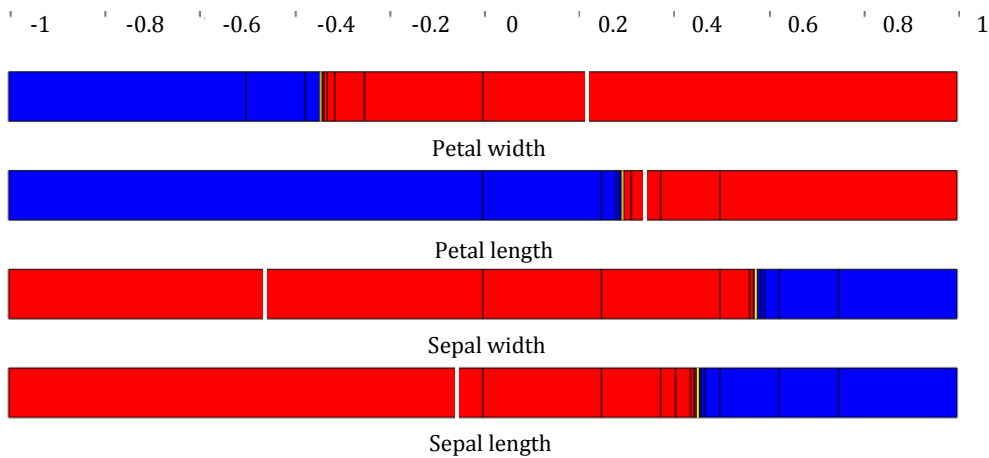}
	\caption{White traces indicate position of the feature values for a pattern in ({\em{Iris-versicolor}}) which is incorrectly classified in ({\em{Iris-virginica}}) class.}
	\label{section4:fig3}
\end{figure}

Given that this pattern is classified in the {\em{Iris-virginica}} class it is possible to check the relevance of each feature to this prediction. In fact, Figure \ref{section4:fig3}, below, shows in details the relevance of the features for this pattern now considered in the {\em{Iris-virginica}} class, that is, as predicted by MLP-2. The values of the features are noted as in the case of Figure \ref{section4:fig2} with white vertical traces in the red colored areas of the features i.e. inside the ranges of values contributing to predict the {\em{Iris-virginica}} class. 

\begin{figure}
\centering

\subfloat{\includegraphics[clip, trim=0.75cm 15.0cm 0.75cm 3.0cm, width=0.95\textwidth]{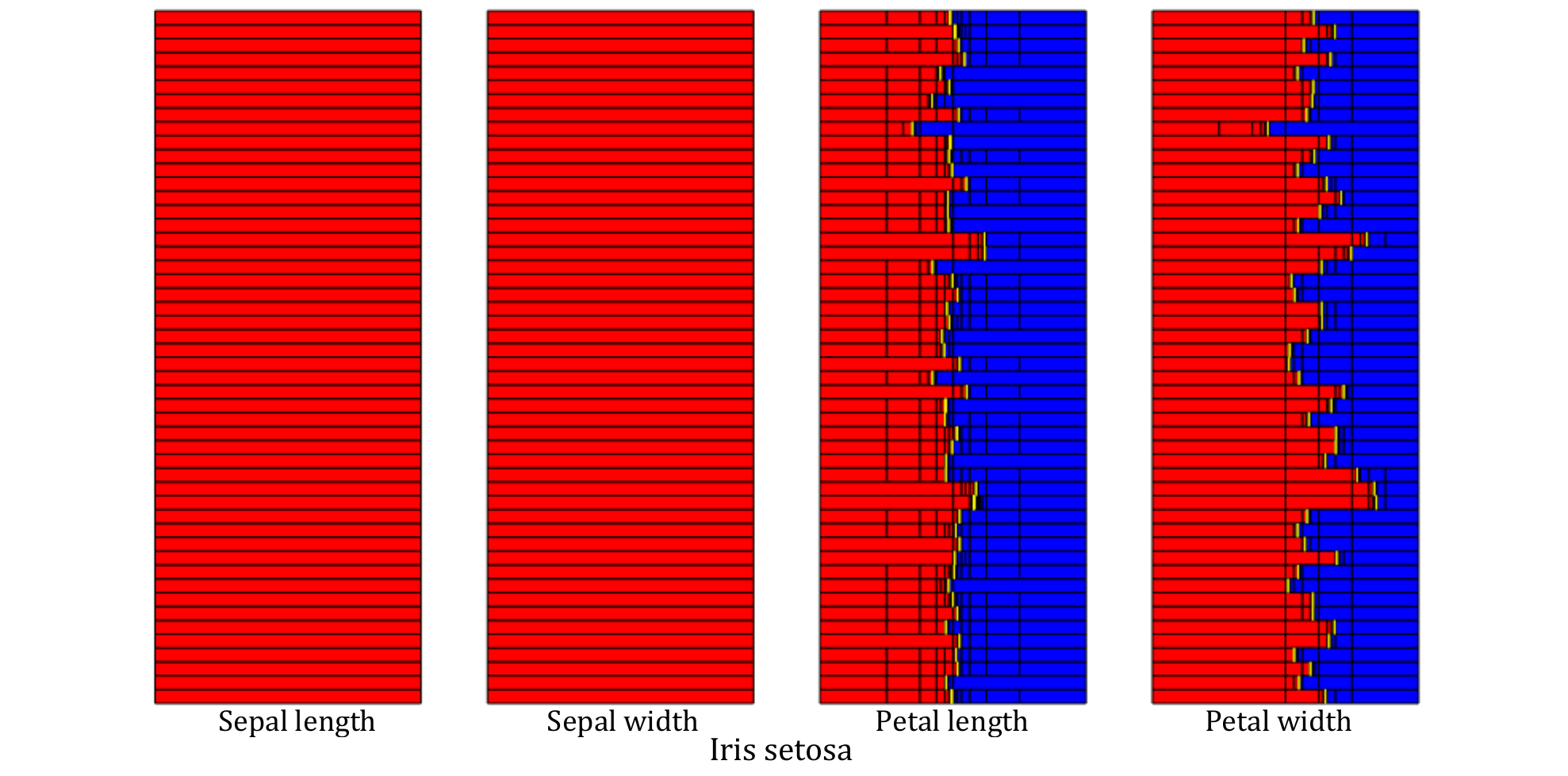}
\centering
\label{section4:fig4:1}}
\vspace{-5mm}
\vfill

\subfloat{\includegraphics[clip, trim=0.75cm 15.0cm 0.75cm 3.0cm, width=0.95\textwidth]{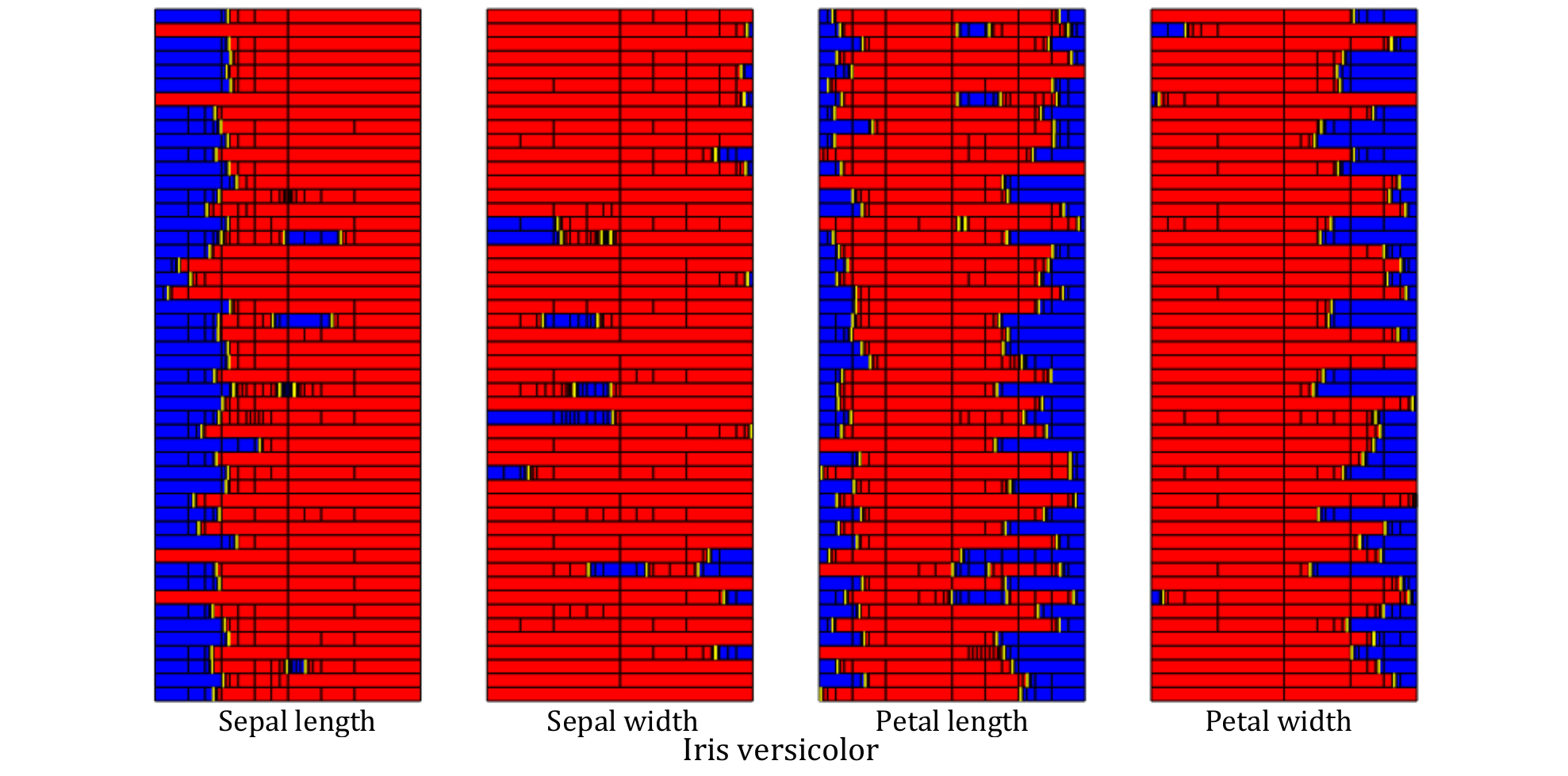}
\centering
\label{section4:fig4:2}}
\vspace{-5mm}
\vfill

\subfloat{\includegraphics[clip, trim=0.75cm 15.0cm 0.75cm 3.0cm, width=0.95\textwidth]{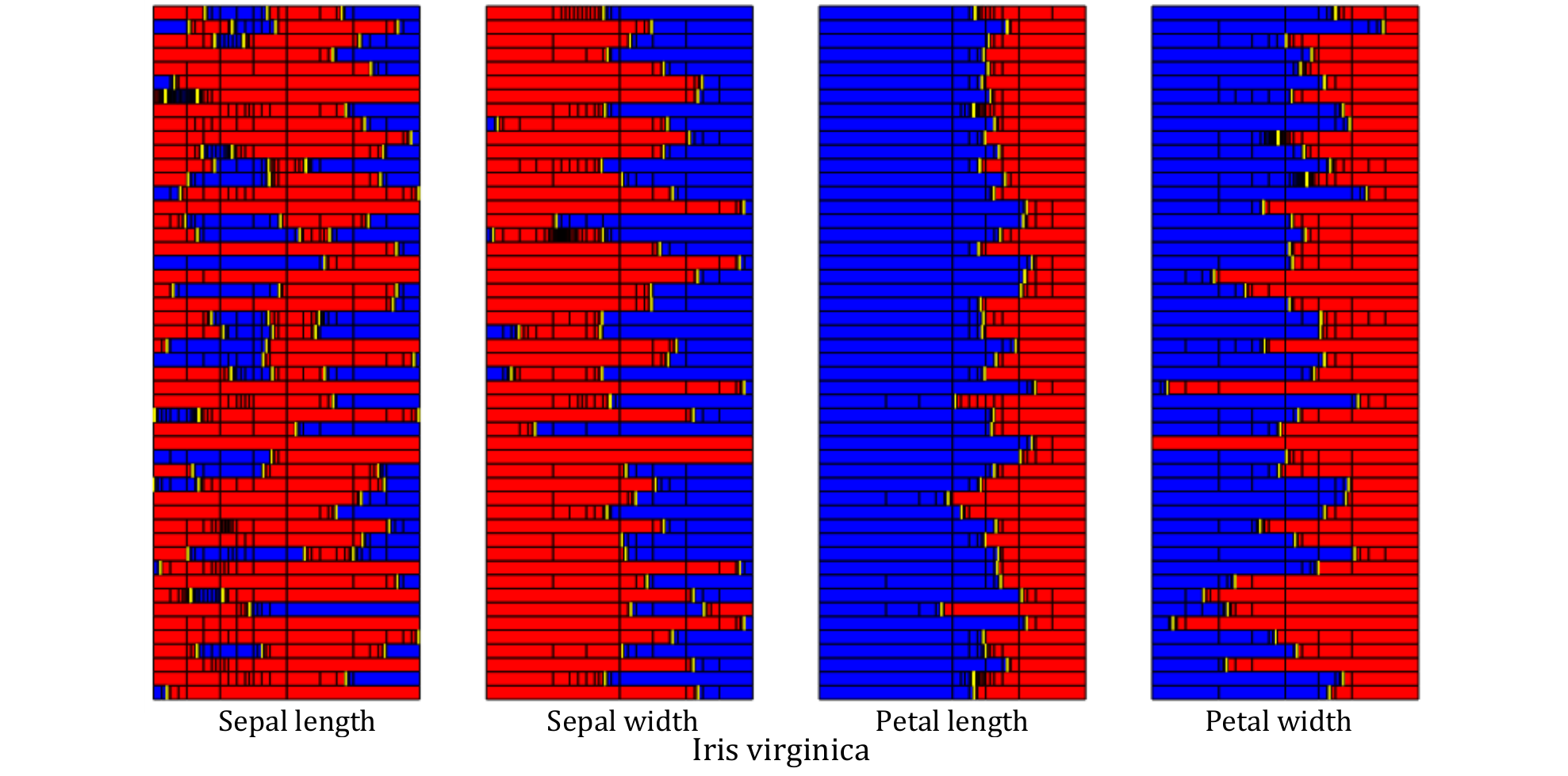}
\label{section4:fig4:3}}
\centering
\caption{{Relevance Map for features of MLP-8. }}
\label{section4:fig4}
\vspace{-7.5 mm}
\end{figure}

In addition to the above, Figure \ref{section4:fig4} depicts the same information, as Figure \ref{section4:fig1}, but in this case the MLP has 8 hidden units (MLP-8). Here, one is able to notice that the classification model implemented by MLP-8 is completely different from the model implemented by MLP-2. For instance features {\em{Sepal length}} and {\em{Sepal width}} have zero relevance for predicting any pattern of the {\em{Iris-setosa}} class. On the other hand these features seem to have a higher relevance for predicting patterns of the {\em{Iris-virginica}} class, while this is not the case in the respective case of MLP-2, see Figure \ref{section4:fig1}. So, explanations that can be formulated for these predictions are completely different and they depend, absolutely, on the classification model implemented by the classifier.

Concerning Figure \ref{section4:fig4} it is worth noting the fragmentation of features {\em{Sepal length}} and {\em{Sepal width}} for the patterns of the  {\em{Iris-virginica}} class. An additional characteristic in this figure concerns the small areas appearing as black spots. These areas result from contiguous bisections applied by the SIVIA algorithm as in the center of these areas there are isolated values of the corresponding feature contributing to predict the specific class. 

\begin{figure}
\centering
\subfloat[Relevance Map for Sepal length contribution to all classes.]{\includegraphics[clip, trim=0.1cm 12.5cm 0.1cm 0.75cm, width=1.00\textwidth]{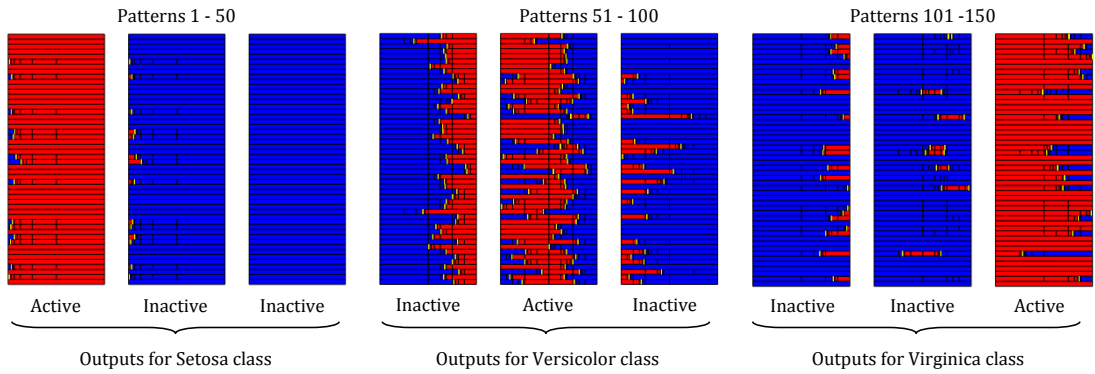}
\label{section4:fig5:1}}
%
\vfill
\centering
\subfloat[Relevance Map for Sepal width contribution to all classes.]{\includegraphics[clip, trim=0.1cm 12.5cm 0.1cm 0.75cm, width=1.00\textwidth]{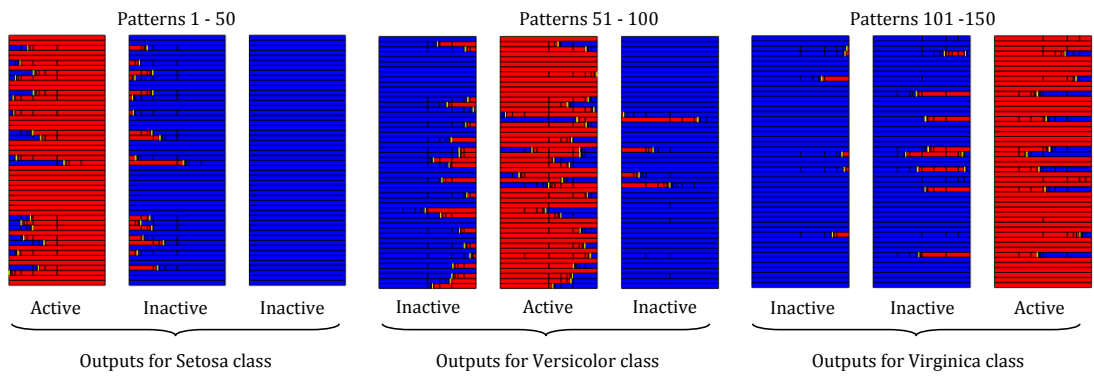}
\label{section4:fig5:2}}
%
\vfill
\centering
\subfloat[Relevance Map for Petal length contribution to all classes.]{\includegraphics[clip, trim=0.1cm 12.5cm 0.1cm 0.75cm, width=1.00\textwidth]{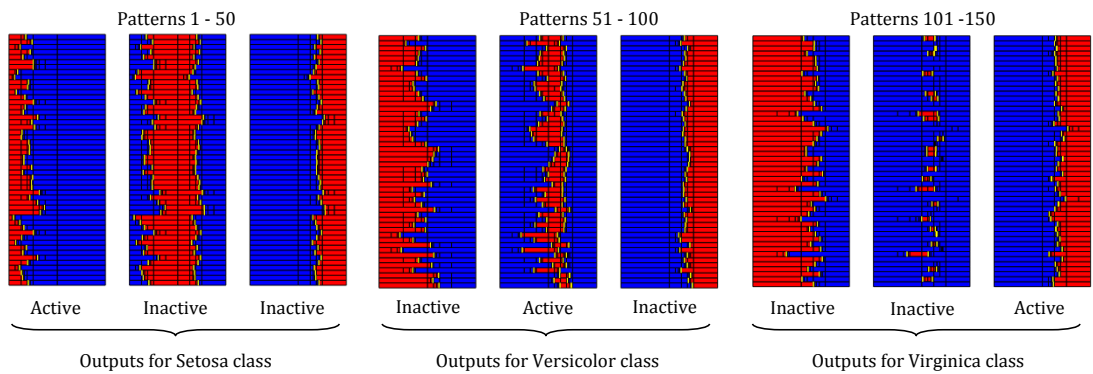}
\label{section4:fig5:3}}
%
\vfill
\centering
\subfloat[Relevance Map for Petal width contribution to all classes.]{\includegraphics[clip, trim=0.1cm 12.5cm 0.1cm 0.75cm, width=1.00\textwidth]{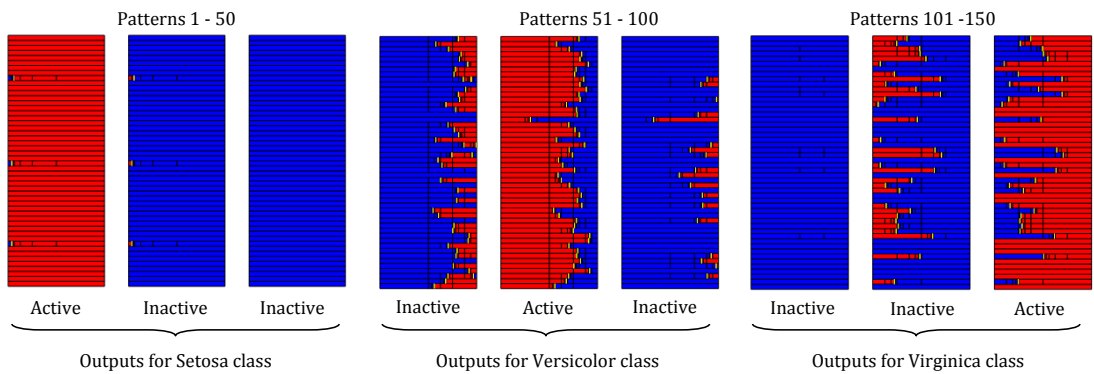}
\label{section4:fig5:4}}
\vfill
\caption{{Relevance Map for all features. Contribution to both active and inactive outputs. }}
\label{section4:fig5}
%
\end{figure}

Figure \ref{section4:fig5} and its sub-figures summarize the relevance of each feature, for each pattern in every class. The Relevance Maps portray in each case the contribution of a feature not only to the active output related to a class but also to the outputs that need to be kept inactive for this class, according to the $1$-of-$M$ codification. Here, while the intervals for the active outputs are all taken to be $[0.8,1]$ the intervals for inactive outputs are considered to be $[0,0.2]$. In this figure one is able to notice that zero relevance is the value for those features that do not contribute at all to some MLP outputs. Note that this happens only for features and inactive outputs. This remark supports the comment done on the value of a feature when ${\mu({\hat{\textrm{x}}}_{k})}=0$ as in Formula \ref{relevance_formula_2} and so the relevance value for the features in these cases is set to zero and not to $1$. 

In all the above experiments with the Iris dataset, the output intervals $[o_{j}-\beta_{j},o_{j}+\beta_{j}]$ are taken to be fairly wide with radii $\beta_{j}=0.2$. This choice aimed to derive larger areas in the range of values of the features in order to provide figures which clearly display the result of the set membership estimation procedure. On the other hand these results reveal the contribution of the features to surely predict specific classes for  the patterns. 

\begin{figure}
\centering
\subfloat{\includegraphics[clip, trim=0.75cm 15.0cm 0.75cm 3.0cm, width=0.95\textwidth]{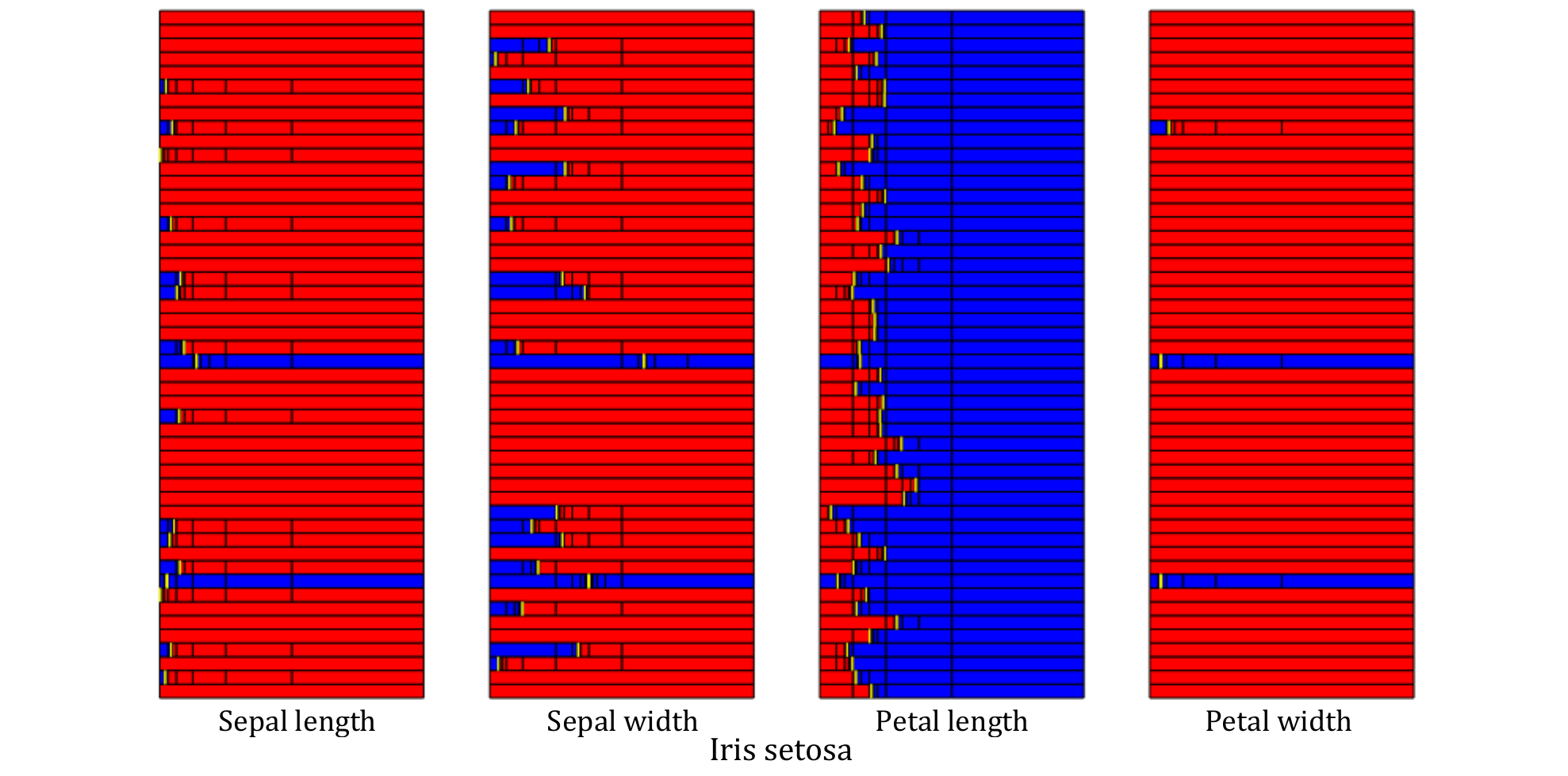}
\centering
\label{section4:fig6:1}}
\vspace{-5mm}
\vfill

\subfloat{\includegraphics[clip, trim=0.75cm 15.0cm 0.75cm 3.0cm, width=0.95\textwidth]{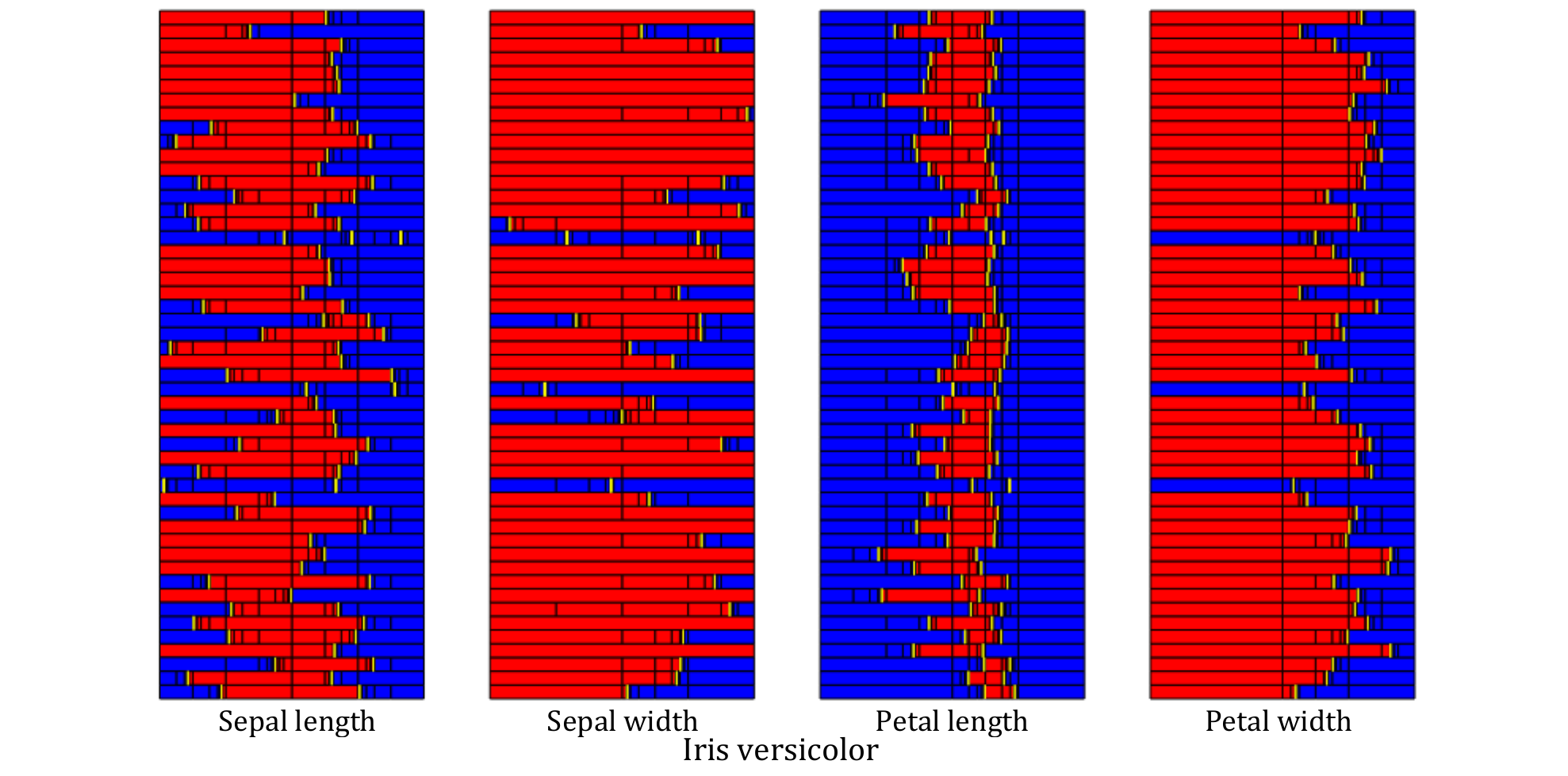}
\centering
\label{section4:fig6:2}}
\vspace{-5mm}
\vfill

\subfloat{\includegraphics[clip, trim=0.75cm 15.0cm 0.75cm 3.0cm, width=0.95\textwidth]{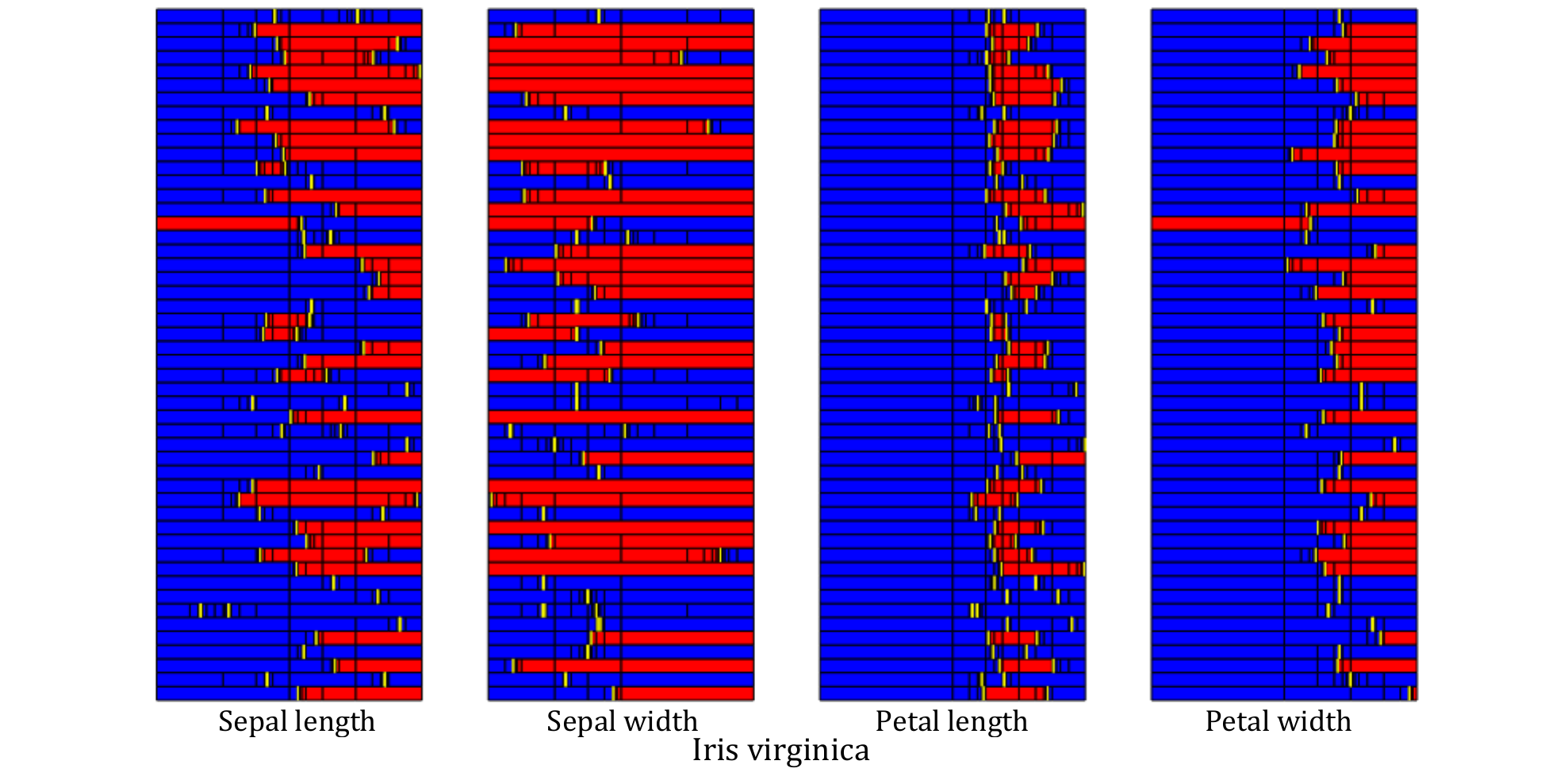}
\label{section4:fig6:3}}
\centering
\caption{\small{Relevance Maps as in Figure \ref{section4:fig1}. Estimation of narrow output intervals (radii are $\beta_{j}=1e-04$) }}
\label{section4:fig6}
\vspace{-7.5 mm}
\end{figure}

From another point of view one might be interested in the feature relevance concerning predictions of the MLP that are as punctual as possible. In such a case one should define a ``sufficiently'' narrow interval $[o_{j}-\beta_{j},o_{j}+\beta_{j}]$ i.e. set a ``sufficiently'' small value to the radius parameter $\beta_{j}$. While the characterization ``sufficiently'' constitutes a heuristic term it is unavoidable to remark that it's up to the user and the explanation information he wants to have to specify a value for $\beta_{j}$. A meaningful example for this case is given in Figure \ref{section4:fig6} where the value set for $\beta_{j}=1e-04$ and in consequence the value for the resolution parameter $\varepsilon$ of SIVIA is set to $1e-06$. 

The obvious comment that can be done regarding the results portrayed by Figure \ref{section4:fig6} is that the narrower the output interval $[o_{j}-\beta_{j},o_{j}+\beta_{j}]$ the narrower the range of values of the features in the input space and so the higher its relevance. However, the comment that seems to be instructive here concerns the blue colored areas in the range of certain features for some specific patterns disrupted by thin yellow lines denoting some interval of extremely small range or even some single value for these features. This result supports the comment done on the relevance of a feature when ${\mu({\hat{\textrm{x}}}_{k})} \approxeq 0$ in formula \ref{relevance_formula_2} for which we consider ${\mathcal{R}}_{\mathcal{F},{\mathbf{x}},k} = 1$.    

The last experiment that merits to be presented, here, concerns application of the proposed approach for discovering the relevance of those features which are intrinsically irrelevant to the classification problem at hand. To this end we created an artificial ``Iris dataset'' by adding to the original one two artificial features with random values uniformly distributed in the interval $[-1,1]$. While preserving the labels of the original dataset, we used an MLP-2 with exactly the same architecture (hidden layer, nodes and activation functions) as the MLP reported in the beginning of this subsection. Training this network with the same algorithm and using the same parameters gave the same results concerning minimization of the loss function and classification of the training patterns. 

\begin{figure}
\centering
\subfloat{\includegraphics[clip, trim=1.75cm 17.0cm 0.75cm 3.0cm, width=0.95\textwidth]{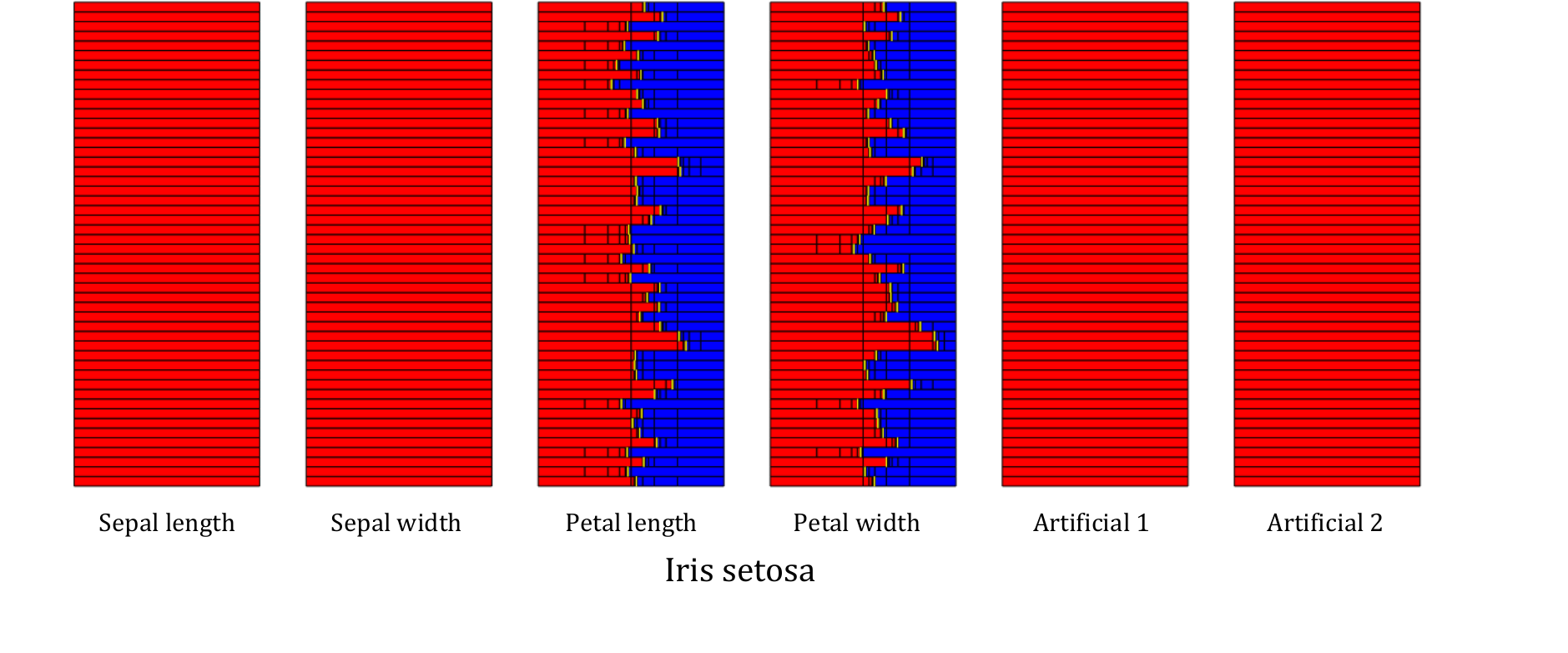}
\centering
\label{section4:fig7:1}}
\vspace{-5mm}
\vfill

\subfloat{\includegraphics[clip, trim=1.75cm 17.0cm 0.75cm 3.0cm, width=0.95\textwidth]{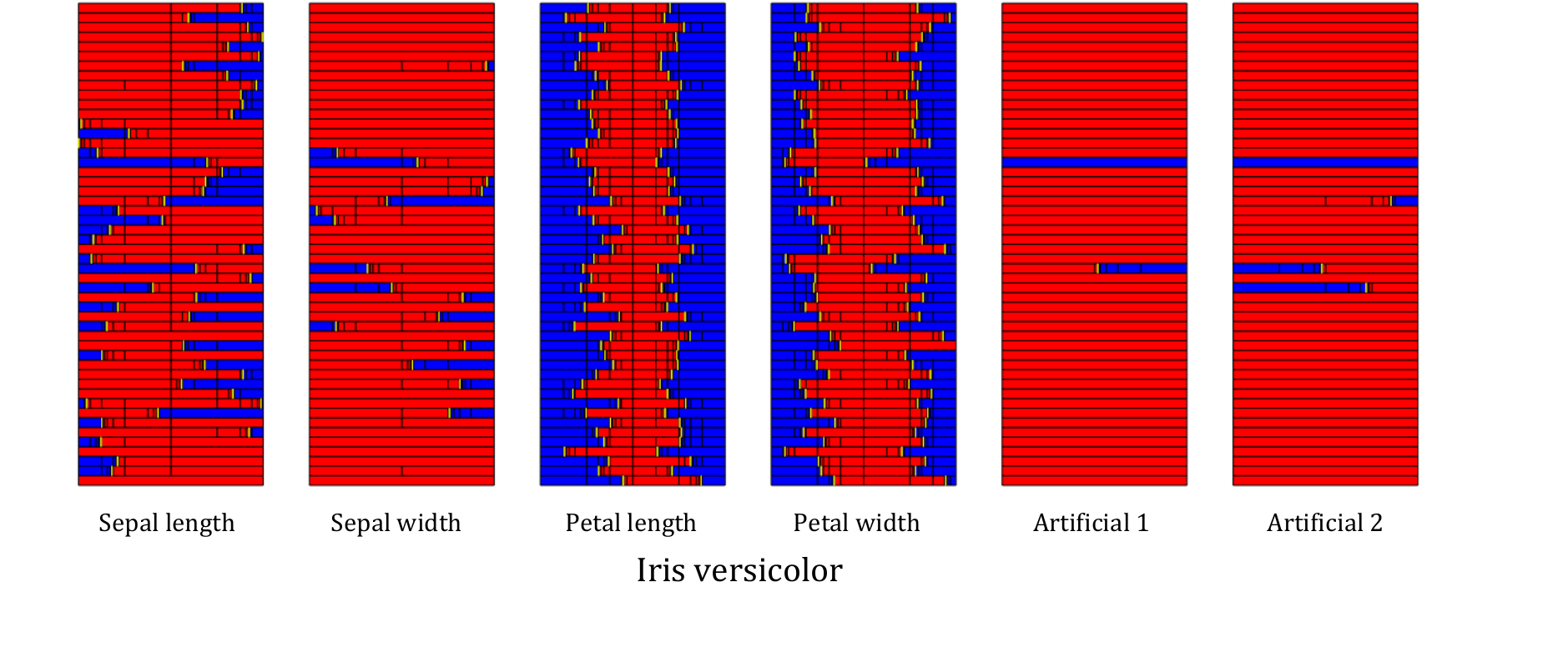}
\centering
\label{section4:fig7:2}}
\vspace{-5mm}
\vfill

\subfloat{\includegraphics[clip, trim=1.75cm 17.0cm 0.75cm 3.0cm, width=0.95\textwidth]{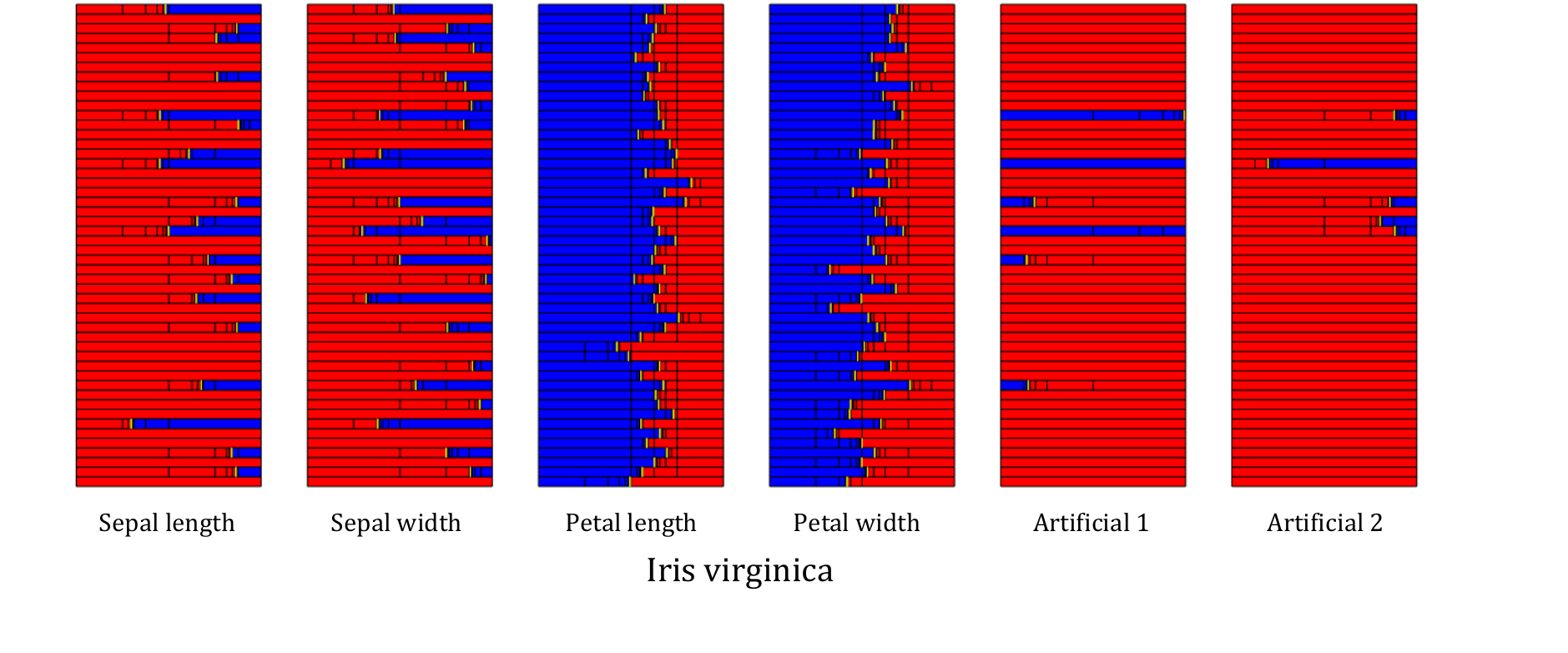}
\label{section4:fig7:3}}
\centering
\caption{{Relevance Maps with the artificial features of the Iris dataset.}}
\label{section4:fig7}
\vspace{-7.5 mm}
\end{figure}

Figure \ref{section4:fig7} and its sub-figures summarize the relevance of each feature, including the artificial ones, for each pattern in every class. In this figure one is able to notice that the artificial features do not contribute to the MLP outputs. For the few patterns that the artificial features appear to have some contribution this seems to be a kind of ``random activity'', possibly due to the fact that these feature values lie in the proximity of some decision surface. 

\subsection{Experiments on the MNIST Dataset}

MNIST is the classic dataset with images of handwritten digits that has served as the basis for benchmarking classification algorithms and still remains a reliable resource for researchers as new machine learning techniques emerge. It comprises a training set of $60,000$ examples, and a test set of $10,000$ examples of digits that have been size-normalized and centered in a fixed-size image of $784 (=28 \times 28)$ pixels containing gray levels as a result of the anti-aliasing technique used by the normalization algorithm. 

Typically, classification of handwritten characters is performed using convolutional and deep learning neural architectures. However, as in this work we are interested in studying the effectiveness of set membership estimation in interpreting the decisions of shallow neural classifiers, we used an MLP having one hidden layer with $100$ neurons using the hyperbolic tangent activation function. The network is fully connected and comprises $10$ output neurons, one for each class of digits. Each output neuron provides a prediction for the class of a $28 \times 28$ image using the {\em{softmax}} activation function. The class for the $0$ digit is coded as class number $10$. The network is trained to minimize the {\em{cross entropy}} loss function using the {\em{scaled conjugate}} gradient descent procedure. During training validation was performed using the 20\% of the training set. Finally, note that all images used both for training and test were scaled in the interval $[-1,1]$.

After training classification performance of the MLP was tested for the test images of the dataset and it was found that $96.90$\% of these images were successfully classified. In the sequel, the classification decision of the MLP was investigated for a number of selected test images using the proposed approach. To this end each image tested is considered to be a pattern composed of $784$ features, one per pixel of the $28 \times 28$ pixels of the image. So, set membership estimation, using SIVIA, was applied for each one of the $784$ features with the following considerations. We suppose that for a successfully classified test image representing the $j$-th digit, the value of the $j$-th {\em{softmax}} output node is $v_{j}$. Then the output interval of interest was set to $[v_{j},1]$. In addition the value of the parameter $\varepsilon$ of SIVIA was set to ${v_{j}}*{10}^{-1}$ in order to achieve a ``good'' resolution of the intervals obtained without having significant burden of the computational process. For instance if $v_{j}=0.995$ then $\varepsilon$ was set to $0.1e-4$ and if $v_{j}=1$ then the output interval of interest is set to $[0.9999,1]$ and $\varepsilon=0.1e-5$. 
	
\begin{figure}
\begin{tabular}{cccccccc}
\multicolumn{2}{c}{\includegraphics[width=0.70 in]{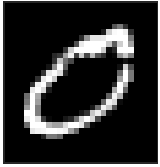}\includegraphics[width=0.70 in]{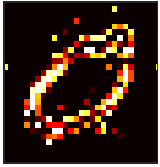}} & \multicolumn{2}{c}{\includegraphics[width=0.70 in]{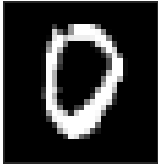}\includegraphics[width=0.70 in]{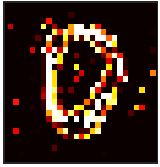}} & \multicolumn{2}{c}{\includegraphics[width=0.70 in]{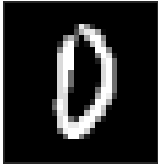}\includegraphics[width=0.70 in]{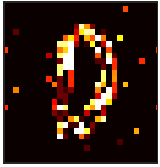}} & \multicolumn{2}{c}{\includegraphics[width=0.70 in]{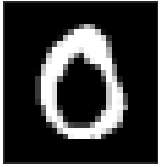}\includegraphics[width=0.70 in]{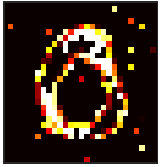}}  \\

\multicolumn{2}{c}{\includegraphics[width=0.70 in]{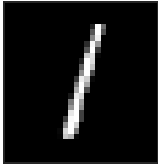}\includegraphics[width=0.70 in]{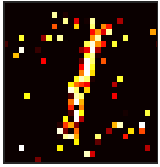}} & \multicolumn{2}{c}{\includegraphics[width=0.70 in]{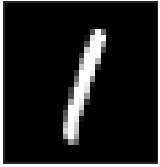}\includegraphics[width=0.70 in]{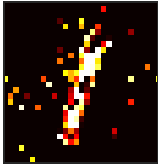}} & \multicolumn{2}{c}{\includegraphics[width=0.70 in]{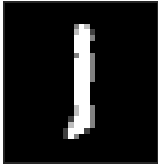}\includegraphics[width=0.70 in]{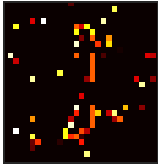}} & \multicolumn{2}{c}{\includegraphics[width=0.70 in]{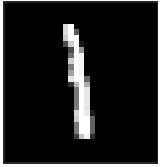}\includegraphics[width=0.70 in]{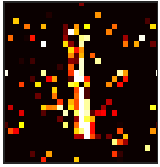}}  \\

\multicolumn{2}{c}{\includegraphics[width=0.70 in]{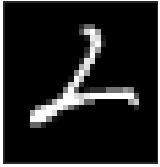}\includegraphics[width=0.70 in]{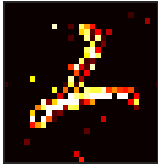}} & \multicolumn{2}{c}{\includegraphics[width=0.70 in]{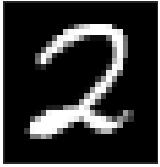}\includegraphics[width=0.70 in]{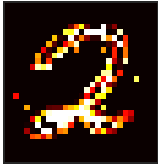}} & \multicolumn{2}{c}{\includegraphics[width=0.70 in]{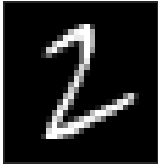}\includegraphics[width=0.70 in]{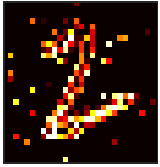}} & \multicolumn{2}{c}{\includegraphics[width=0.70 in]{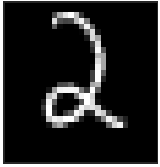}\includegraphics[width=0.70 in]{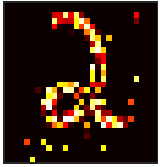}}  \\

\multicolumn{2}{c}{\includegraphics[width=0.70 in]{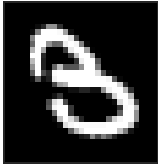}\includegraphics[width=0.70 in]{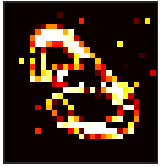}} & \multicolumn{2}{c}{\includegraphics[width=0.70 in]{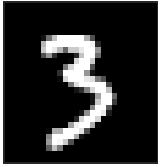}\includegraphics[width=0.70 in]{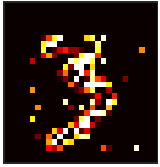}} & \multicolumn{2}{c}{\includegraphics[width=0.70 in]{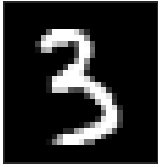}\includegraphics[width=0.70 in]{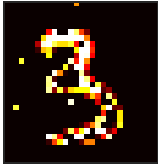}} & \multicolumn{2}{c}{\includegraphics[width=0.70 in]{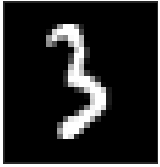}\includegraphics[width=0.70 in]{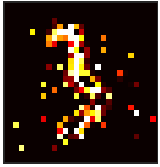}}  \\

\multicolumn{2}{c}{\includegraphics[width=0.70 in]{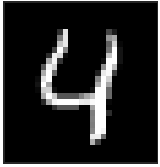}\includegraphics[width=0.70 in]{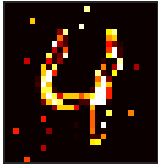}} & \multicolumn{2}{c}{\includegraphics[width=0.70 in]{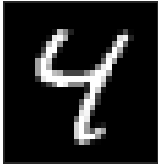}\includegraphics[width=0.70 in]{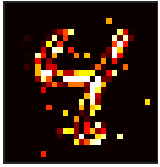}} & \multicolumn{2}{c}{\includegraphics[width=0.70 in]{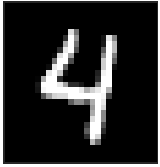}\includegraphics[width=0.70 in]{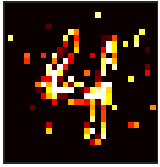}} & \multicolumn{2}{c}{\includegraphics[width=0.70 in]{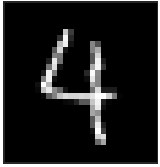}\includegraphics[width=0.70 in]{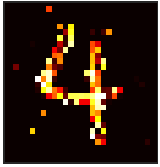}}  \\

\multicolumn{2}{c}{\includegraphics[width=0.70 in]{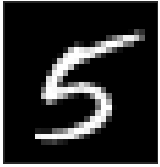}\includegraphics[width=0.70 in]{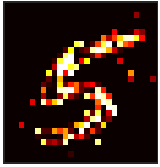}} & \multicolumn{2}{c}{\includegraphics[width=0.70 in]{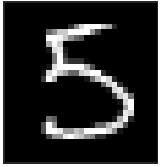}\includegraphics[width=0.70 in]{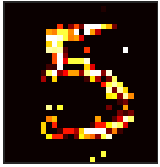}} & \multicolumn{2}{c}{\includegraphics[width=0.70 in]{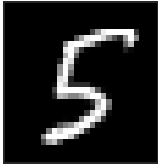}\includegraphics[width=0.70 in]{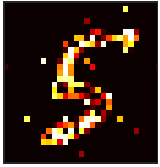}} & \multicolumn{2}{c}{\includegraphics[width=0.70 in]{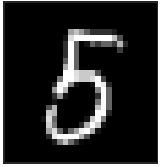}\includegraphics[width=0.70 in]{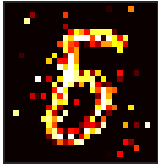}}  \\

\multicolumn{2}{c}{\includegraphics[width=0.70 in]{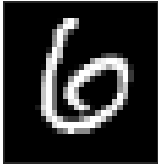}\includegraphics[width=0.70 in]{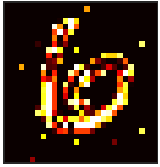}} & \multicolumn{2}{c}{\includegraphics[width=0.70 in]{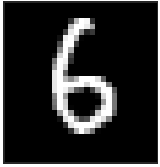}\includegraphics[width=0.70 in]{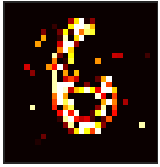}} & \multicolumn{2}{c}{\includegraphics[width=0.70 in]{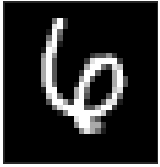}\includegraphics[width=0.70 in]{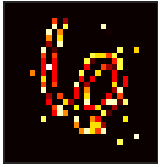}} & \multicolumn{2}{c}{\includegraphics[width=0.70 in]{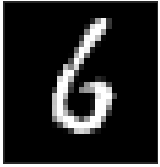}\includegraphics[width=0.70 in]{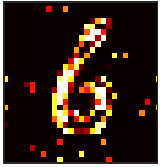}}  \\

\multicolumn{2}{c}{\includegraphics[width=0.70 in]{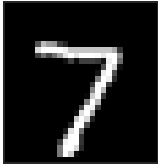}\includegraphics[width=0.70 in]{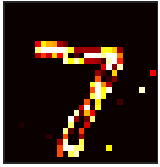}} & \multicolumn{2}{c}{\includegraphics[width=0.70 in]{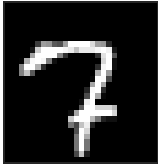}\includegraphics[width=0.70 in]{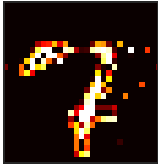}} & \multicolumn{2}{c}{\includegraphics[width=0.70 in]{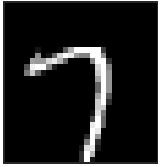}\includegraphics[width=0.70 in]{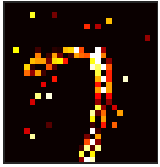}} & \multicolumn{2}{c}{\includegraphics[width=0.70 in]{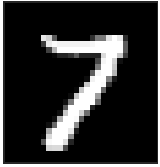}\includegraphics[width=0.70 in]{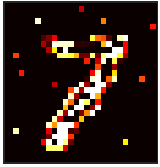}}  \\

\multicolumn{2}{c}{\includegraphics[width=0.70 in]{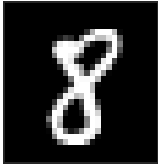}\includegraphics[width=0.70 in]{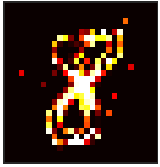}} & \multicolumn{2}{c}{\includegraphics[width=0.70 in]{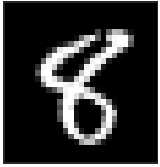}\includegraphics[width=0.70 in]{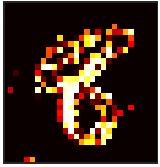}} & \multicolumn{2}{c}{\includegraphics[width=0.70 in]{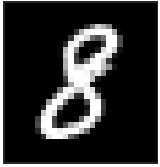}\includegraphics[width=0.70 in]{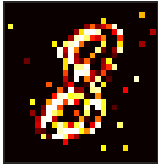}} & \multicolumn{2}{c}{\includegraphics[width=0.70 in]{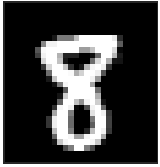}\includegraphics[width=0.70 in]{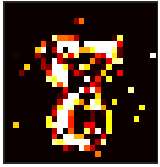}}  \\

\multicolumn{2}{c}{\includegraphics[width=0.70 in]{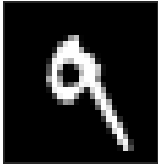}\includegraphics[width=0.70 in]{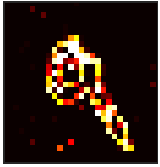}} & \multicolumn{2}{c}{\includegraphics[width=0.70 in]{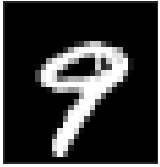}\includegraphics[width=0.70 in]{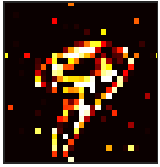}} & \multicolumn{2}{c}{\includegraphics[width=0.70 in]{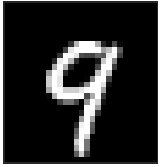}\includegraphics[width=0.70 in]{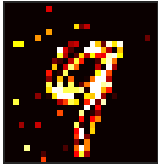}} & \multicolumn{2}{c}{\includegraphics[width=0.70 in]{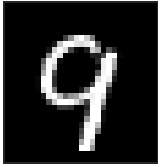}\includegraphics[width=0.70 in]{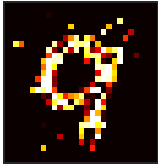}}  \\

\end{tabular}
\caption{Performance of the proposed approach for a number of images from the MNIST dataset}
\label{section4:figure8}
\end{figure}

For each correctly classified image the prediction of the MLP was analyzed and the relevance computed using formula (\ref{relevance_formula_2}) was imprinted on a $28 \times 28$ pixels image using the ``hot'' color-map of MATLAB {(https://www.mathworks.com/help/matlab/ref/colormap{\_}hot.png)}. In this color palette white stands for the highest relevance while black represents zero relevance. Figure \ref{section4:figure8} displays a number of representative examples of the application of this process. So, for each original gray scale image the adjacent image portrays the degree to which each pixel contributes to the prediction made by the neural classifier. One may easily conclude that a successful prediction of the classifier is based upon the pixel values at the edges of the gray scale figure and/or those pixels constituting its corpus. However, as shown in Figure\ref{section4:figure8} several pixels while located at a certain distance from the figure corpus appear to influence the classification decision of the network. A number of hypotheses can be made on this subject but since the objective of this paper is to demonstrate the potential of the proposed approach and this experiment uses an MLP, i.e. a shallow neural network which is not known to be particularly appropriate for this benchmark, we will not comment this matter here, leaving it as an open question for our future work.

\section{Conclusion and Future Work}      

In this preprint we presented a new approach to discovering feature relevance and explaining decisions of neural classifiers. The proposed approach builds on set membership estimation which has proven to be effective in parameter estimation and system identification tasks, especially in the domain of control systems. Furthermore, computation of the feasible set relies on interval analysis techniques which constitute a sound mathematical background ensuring the reliability of the results obtained. The proposed approach seems to act similarly to existing techniques which are based on sensitivity analysis, on relevance propagation (LRP), or on backpropagation of a black-box model prediction to the input feature like in DEEPLift, etc. However, despite this similarity in terms of the targeted elements of the classifier and their characterization, it is significantly different as it does not rely on heuristic assumptions but it is entirely based on the classifier's implementation of the classification function. Moreover, the algorithm implemented makes no use of gradient computation while the analysis of the feature relevance it performs is very similar to known sensitivity analysis methods. The algorithm in use is particularly efficient in terms of computational time despite the fact that it explores exhaustively the input range of each feature. Further testing with known benchmarks and comparison with related techniques will be part of our immediate work plans.

The examples obtained on two well known benchmarks, Fisher's Iris classification dataset and the MNIST dataset of handwritten character recognition, assess the validity of the proposed approach on shallow neural classifiers and underline the need to further investigate its use on deep learning architectures. Moreover, it would be interesting to check the relation of this approach with different saliency methods found in the literature. Other possible directions include the production of a consistent interpretation of the proposed feature relevance maps in terms of more a interpretable model (e.g. decision tree) and possibly the extraction of rules learned by the classifier.

\medskip
\bibliographystyle{unsrtnat}
\bibliography{References}

\end{document}